%% file: main.tex
\documentclass[
  a4paper,        
  techreport,     
]{art}

\usepackage{subscript}
\usepackage{graphicx}
\usepackage{wrapfig}
\usepackage{microtype}

\usepackage{amsopn}

\usepackage[dvipsnames]{xcolor}
\usepackage{enumitem}
\usepackage{fontawesome} 
\usepackage{fvextra}  
\usepackage{tablefootnote} 
\usepackage[ruled,vlined]{algorithm2e} 
\usepackage{minted} 
\usepackage{natbib}
\usepackage{hyperref}

\usepackage{tcolorbox}
\usepackage{float}
\usepackage{booktabs}
\usepackage{adjustbox}
\usepackage{pdflscape}

\usepackage[utf8]{inputenc}
\usepackage{textcomp}

\usepackage{pifont}  

\usepackage[normalem]{ulem}

\usepackage{paracol}

\usepackage{tabularx}   
\newcolumntype{Y}{>{\raggedright\arraybackslash}X}
\newcolumntype{C}{>{\centering\arraybackslash}X} 
\tcbuselibrary{skins,breakable}

\definecolor{RuleGray}{gray}{0.82}
\definecolor{TitleGray}{gray}{0.20}
\definecolor{FillGray}{gray}{0.985}

\tcbset{
  enhanced,
  colback=FillGray,
  colframe=RuleGray,
  coltitle=TitleGray,
  fonttitle=\bfseries\scshape,
  arc=1.6mm,
  boxsep=1.0ex,
  left=1.1ex,right=1.1ex,top=1.0ex,bottom=1.0ex,
  borderline horizontal={0.6pt}{0pt}{RuleGray},
}

\newtcolorbox{ProblemBox}{
  title=Problem,
  colback=white,
  borderline horizontal={0.8pt}{0pt}{RuleGray},
  breakable,  
}

\newtcolorbox{RespBox}[1]{
  title=#1,
  breakable,  
  pad at break*=2mm,
  skin=enhanced,
}

\definecolor{boxbg}{RGB}{248, 249, 250}
\definecolor{bordercol}{RGB}{59, 130, 246}

\definecolor{bg}{rgb}{0.95,0.95,0.95}

\usepackage{titlesec}

\definecolor{principle1}{HTML}{E3F2FD}  
\definecolor{principle2}{HTML}{E8F5E9}  
\definecolor{principle3}{HTML}{FFF3E0}  
\definecolor{principle4}{HTML}{F3E5F5}  
\definecolor{baseline}{HTML}{FCE4EC}   

\tcbset{
  sharp corners,
  boxrule=0.5pt,
  colframe=black,
  fonttitle=\bfseries,
  title filled,
}
\input{metadata.tex}
\input{macros.tex}

\begin{document}

\newgeometry{left=2.5cm,right=2.5cm,bottom=2cm}

\maketitle

\begin{figure*}[htbp] 
    \centering
    \includegraphics[width=\textwidth,height=0.35\textheight,keepaspectratio]{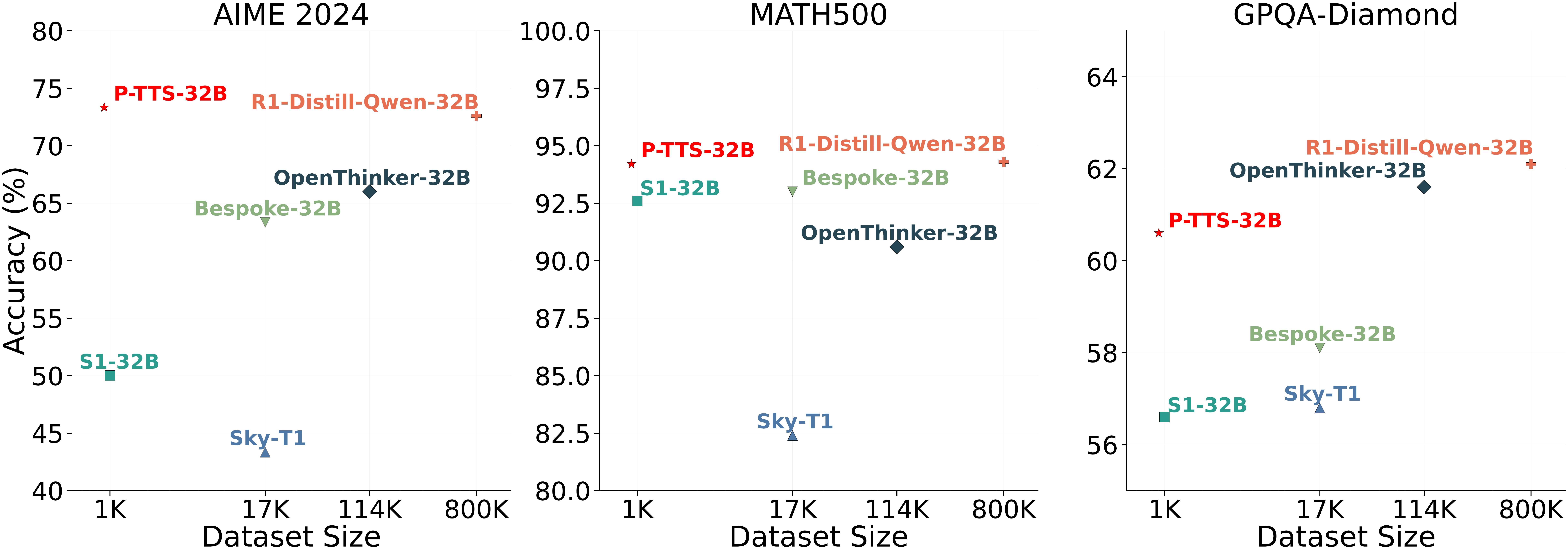}
    \caption{
        \textbf{Comparison of 32B-scale models on AIME 2024 (left), MATH500 (middle), and GPQA Diamond (right).} 
        Model performance on AIME 2024, MATH500, and GPQA-Diamond benchmarks as a function of dataset size. 
        Each point represents a different model, with our \texttt{P-TTS-32B} (red star) showing competitive performance from a significantly smaller dataset. 
        The x-axis scale highlights the differences in training data sizes across models.
    }
    \label{fig:main_results}
\end{figure*}
\section{Introduction}

Large language models (LLMs)~\citep{radford2018improving,achiam2023gpt,team2023gemini,anthropic2025claude3_7_sonnet} attain strong deductive and quantitative reasoning once equipped with curated chains of thought (CoT)~\citep{wei2022chain} or tool-augmented exemplars~\citep{ma2024sciagent}. However, constructing thousand-scale reasoning corpora is costly: it requires prompt engineering, human verification of multi-step solutions, and continuous refresh to track dataset shifts {\em in the wild}. Moreover, static large-shot prompts are brittle—fixed exemplars can inadvertently bias the model toward spurious solution templates or fail under domain shift, limiting generalization despite high in-domain scores.
Prior work has largely scaled pre-/post-training time data (pre-training, instruction tuning, supervised CoT) or model size, while inference-time strategies typically vary only in decoding parameters (temperature, sampling) or rerank multiple outputs from a single prompt. The combinatorial space of which exemplars to show, how to order them, and how to perturb them remains mostly unexploited. We argue that the prompt itself is a stochastic control knob whose systematic scaling at test time can simulate the effect of large reasoning datasets, without actually collecting them.

In this work, we propose \texttt{\uline{P}rompting \uline{T}est-\uline{T}ime \uline{S}caling (\texttt{P-TTS})} as an LLM data augmentation method: given a compact seed pool of just 90 high-quality math reasoning exemplars, we algorithmically expand the prompt space at inference by 1) exemplar subsampling under diversity constraints using various principled instructions~\citep{bsharat2023principled}, 2) ordering perturbations that modulate inductive biases (recency, primacy), and 3) pseudo-sampling of paraphrased rationales and solution skeletons via the model itself. Each seed exemplar question is paired with a prompt ensemble, a set of independently constructed prompt contexts, whose answers are collected. This converts test-time prompting into a scalable augmentation pipeline.

\noindent{\bf Why can 90 beat 1K?} The key insight of our framework is that, conventionally, a fixed 1K-shot sample or prompt provides only one (or a few) points in the prompt-combinatorial space, whereas our \texttt{P-TTS} explores a far larger manifold of reasoning cues. From a bias–variance perspective, prompt ensembles for the same question reduce variance in reasoning trajectories and increase coverage of latent solution schemas. Information-theoretically, diverse promptings expose the model to a richer set of conditional priors over intermediate steps, effectively approximating a mixture-of-experts CoT without extra training. Empirically, we find that \texttt{P-TTS} surpasses the gap to 1K-shot baselines across reasoning and zero-shot generalization on out-of-domain tasks, and substantially improves robustness on naturally occurring, shifted test distributions.

We verify the effectiveness of \texttt{P-TTS} through two orthogonal dimensions: 1) Semantic/knowledge diversity in CoT responses – measuring how \texttt{P-TTS} expands the coverage of knowledge and concepts. We compare the trade-off between accuracy and knowledge diversity gain across four principled prompting strategies: \texttt{Reward, Correctness, Penalty,} and \texttt{Think}. 2) Language trigram diversity – quantifying lexical and phrasing variety using distinct n-gram ratios and entropy, applied both to full responses and isolated reasoning traces. Results show that reward framing achieves the greatest lexical diversity, reflecting stronger surface-level variation from the base examples. More broadly, we observe synergistic benefits: semantic/knowledge diversity reduces overfitting to narrow reasoning templates and enhances transfer; trigram diversity mitigates lexical/template lock-in and strengthens robustness; ordering perturbations counter position bias slightly; and our model-driven self-augmentation introduces novel but still on-manifold rationales. Finally, scaling curves reveal diminishing returns beyond roughly 6 prompt-augmentations per base question, suggesting this as a practical deployment point.

\noindent{\bf Contributions of this work}. (1) We introduce \texttt{Prompting Test-Time Scaling}, a simple yet effective framework for LLM inference-time reasoning data augmentation. (2) We demonstrate that only 90 seed samples, when leveraged through our \texttt{P-TTS}, can outperform 1K-shot static prompts, reducing curation cost by an order of magnitude. (3) We provide empirical evidence that prompt-space exploration is an underutilized scaling dimension for LLM reasoning. We release code and the augmented exemplar pool to facilitate reproducibility and facilitate rapid transfer to new domains. Collectively, \texttt{P-TTS} reframes test-time prompting from a one-shot design choice into a scalable, stochastic process, unlocking robust reasoning without additional data collection or massive labeled datasets.

\section{Related Work}

\paragraph{LLM Reasoning} The growing availability of frontier autoregressive pretrained large language models that expose long-form rationales (e.g., GPT-o1~\citep{jaech2024openai}, Gemini~\citep{team2023gemini}, Claude 3.7 Sonnet~\citep{anthropic2025claude3_7_sonnet}, DeepSeek-R1~\citep{guo2025deepseek}) has integrated RL or SFT pipelines that \emph{explicitly} incorporate intermediate reasoning. Diffusion LLM~\citep{li2025survey,zhao2025d1} also shows potential in reasoning capability recently. In many open-sourced reasoning LLM settings~\citep{muennighoff2025s1,bespoke_stratos_32b,guha2025openthoughts}, a teacher model is usually prompted to produce chain-of-thought (CoT) explanations for curated inputs, and the resulting (prompt, rationale, answer) triples supervise a student model. For instance, \textsc{Bespoke-Stratos}~\citep{bespoke_stratos_2025} collects teacher-generated explanations from DeepSeek-R1, while the OpenThinker models~\citep{guha2025openthoughts} train on the OpenThoughts corpora built with rationales elicited from teachers such as DeepSeek-R1 or QwQ-32B. These methods demonstrate that model-generated reasoning is a scalable supervision signal, but they typically rely on tens to hundreds of thousands of exemplars and substantial collection costs. Our work targets a complementary regime: we exploit \emph{instructional wrapping} to elicit diverse, high-utility rationales from only 90 seeds, yielding data efficiency competitive with 1K-scale baselines.

\paragraph{Inference-Time Scaling} 
Orthogonal to parameter or dataset scaling~\citep{brown2020language,kaplan2020scaling}, inference-time strategies improve performance without updating model weights. Classic approaches tune decoding hyperparameters (temperature, top-$k$, nucleus sampling) to modulate diversity~\citep{holtzman2019curious,fan2018hierarchical,meister2023locally}, or generate multiple completions from a fixed prompt and aggregate answers, as in few-shot CoT~\citep{wei2022chain} and Self-Consistency~\citep{wang2022self}. Recent “think-more” style methods (e.g., s1/s1.1) adjust the \emph{reasoning budget}—allocating more deliberate tokens at test time—without altering exemplars or performing additional training~\citep{muennighoff2025s1}. We build on this line by treating the \emph{prompt itself} as a first-class scaling axis: \texttt{P-TTS} systematically varies instructional framing to create an ensemble of prompt contexts that can be aggregated at test time or distilled into compact training sets.

\paragraph{Data Augmentation for Reasoning Tasks}
A complementary line synthesizes new samples or rationales to expand training corpora. \textsc{MetaMath}~\citep{yu2023metamath} bootstraps diverse math problems via generate-and-verify loops, while \textsc{ReasoningMix}~\citep{zheng2025learning} composes traces by interleaving steps across tasks. Such methods operate at the \emph{data level} by creating novel items or reasoning sequences. In contrast, \texttt{P-TTS} operates at the \emph{prompt level}: it preserves the original problems but injects controlled diversity through principle-guided wrappers (reward/penalty framing, correctness emphasis, step-by-step cues), which we show can be leveraged both at test time (ensemble prompting) and at training time (SFT over wrapper-elicited rationales).

\paragraph{Principled instructions and prompt engineering}
General-purpose prompting frameworks catalog instruction patterns that improve reliability and adherence~\citep{bsharat2023principled}. Our approach instantiates a small subset of such patterns that are directly compatible with math-style reasoning (reward, penalty, correctness, step-by-step) and formalizes them as deterministic wrap operators. This yields a semantically invariant augmentation space with clear ablation handles (template choice, placement, and paraphrase strength).

\paragraph{Low-Resource Supervision}
Recent efforts show that carefully curated, small datasets can deliver strong reasoning performance. S1~\citep{muennighoff2025s1} and LIMO~\citep{ye2025limo} claim training on $\sim$1K high-quality, challenging prompts as a competitive alternative to massive corpora. Our results complement these findings: with only 90 seeds, \texttt{P-TTS} converts principled prompt-space variation into supervision that matches or surpasses 1K-shot baselines, highlighting prompt-level scaling as a practical lever for low-resource regimes.

\section{Methodology}

\subsection{Overview}  

We propose \textbf{\texttt{Prompting Test-Time Scaling (\texttt{P-TTS})}}, a reasoning-centric data augmentation framework that expands a compact seed set via \emph{instructional wrapping}. Rather than modifying task semantics, P-TTS applies a family of fixed textual wrappers (``principles'')~\cite{bsharat2023principled} to each seed example, producing prompt variants that preserve the original problem while modulating the \emph{instructional framing} (e.g., reward/penalty cues, correctness emphasis, or step-wise guidance). Concretely, a principle $p$ is realized by a template $\tau_p$ that deterministically wraps the raw question $q$, yielding $q^{(p)} = \tau_p \,\Vert\, q$ (string concatenation $\Vert$). The union of the original prompts and their principle-conditioned variants forms a \emph{\texttt{P-TTS} augmented corpus}. Unlike large-scale supervision or domain-specific curation, \texttt{P-TTS} relies on principle-guided prompt reformulation to elicit high-quality reasoning traces from a teacher model, which are then used for supervised fine-tuning (SFT) of a student. In Sec.~\ref{sec:ablations} we show that individual principles already yield measurable gains over the null prompt, and that training on the full \texttt{P-TTS} corpus (from only 90 seeds) can match or exceed models fine-tuned on substantially more data (Table~\ref{tab:s1_comparison}).

\begin{figure*}[t]
  \centering
  \includegraphics[width=1.0\linewidth]{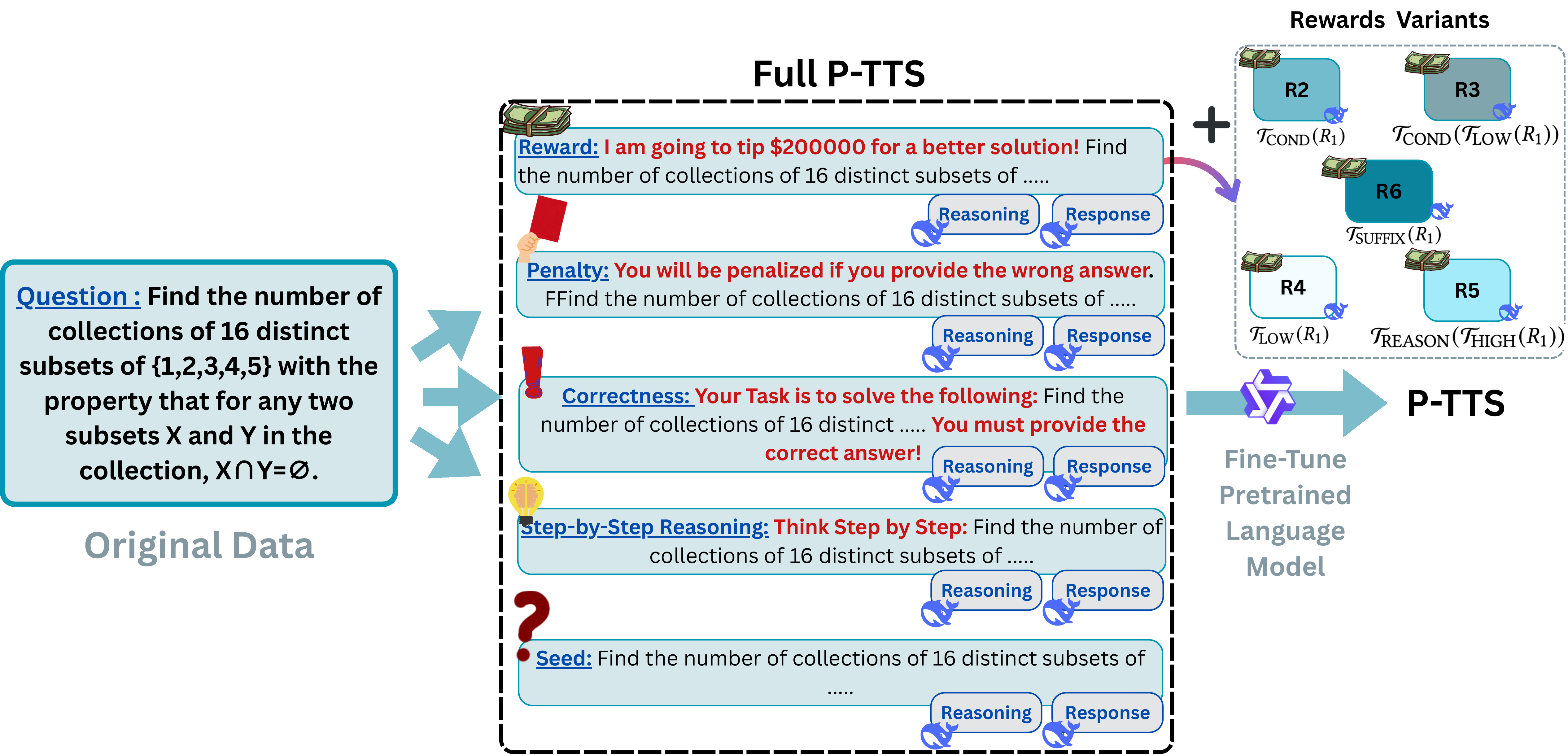}
  \caption{Overview of the \texttt{P-TTS} data augmentation process. Starting from a small set of high-quality math problems (AIME-style), we generate diverse prompt variants through instruction reframing, such as reward-based encouragement, penalty warnings, and step-by-step guidance. These augmented prompts are used to elicit high-quality LLM completions, which are then collected as synthetic reasoning data to fine-tune.
}
  \label{fig:ab}
\end{figure*}

\subsection{Dataset Seed Selection Strategy}
\label{sec:dataset}

We adopt a \emph{seed-based} construction paradigm \citep{zhu2025bare}: a small, vetted set of problems serves as the substrate for systematic instructional variation. Our seeds comprise $N\!=\!90$ problems from AIME 2022–2024, selected for (i) \textbf{reasoning density} across algebra, combinatorics, number theory, geometry, and probability; (ii) \textbf{format and label reliability} (professionally authored items with definitive three-digit answers); and (iii) \textbf{contamination mitigation}---restricting to recent editions reduces overlap with widely scraped, earlier AIME corpora \citep{huang2025thinkbench,achiam2023gpt}. This compact, high-quality set enables controlled scaling via instructional wrappers without compromising semantic fidelity.

\subsection{Selection of Instructional Principles} \label{selection}

We instantiate four core principles
\[
\mathcal{P}_{\mathrm{core}}=\{\texttt{Reward},\ \texttt{Penalty},\ \texttt{Correctness},\ \texttt{StepByStep}\},
\]
chosen for their direct applicability to math reasoning and prior evidence of consistent gains across model families \citep{bsharat2023principled}. Each principle $p$ is bound to a fixed template $\tau_p$ and applied as a wrap operator without modifying $q$’s tokens:
\begin{equation}
q_i^{(p)} = \tau_p \,\Vert\, q_i.
\end{equation}
This guarantees semantic invariance: removing the principled template $\tau_p$ deterministically recovers question $q_i$. Table~\ref{tab:principles_overview} summarizes the templates and induced operators.

\begin{table}[ht]
\centering
\small
\begin{tabularx}{\linewidth}{@{}l Y c@{}}
\toprule
\textbf{Instructional Principle $p$} & \textbf{Template $\tau_p$ (excerpt)} & \textbf{Operator} \\
\midrule
\texttt{Reward}      & \textit{``I am going to tip \$200{,}000 for a better solution!''} & $q^{(\mathrm{rew})}=\tau_{\mathrm{rew}} \,\Vert\, q$ \\
\texttt{Correctness} & \textit{``Your task is\ldots You MUST\ldots''} & $q^{(\mathrm{corr})}=\tau_{\mathrm{corr}} \,\Vert\, q$ \\
\texttt{Penalty}     & \textit{``You will be penalized if you provide the wrong answer.''} & $q^{(\mathrm{pen})}=\tau_{\mathrm{pen}} \,\Vert\, q$ \\
\texttt{StepByStep}  & \textit{``Think step by step.''} & $q^{(\mathrm{step})}=\tau_{\mathrm{step}} \,\Vert\, q$ \\
\bottomrule
\end{tabularx}
\caption{Core \texttt{P-TTS} instructional wrappers. Each operator preserves problem semantics by concatenating a fixed template to the unmodified $q$.}
\label{tab:principles_overview}
\end{table}

\subsection{\texttt{P\textnormal{-}TTS} Dataset Construction}
\label{sec:ptts_data_construction}

Our primary dataset is drawn from the AIME benchmarks (2022--2024) \citep{aime_aops} and consists of \(N=90\) unique problems with gold answers,
\begin{equation}
\mathcal{O}_{\text{seed}}=\{(q_i,a_i\}_{i=1}^{N},
\end{equation}
where \(q_i\) is the original seed question or problem without any additional prompting, and \(a_i\in\{0, 1,\dots,999\}\) is the integer-style ground-truth for the associated question. {{$\mathcal{O}_{\text{seed}}$}} is the seed question-answer pairs.
\paragraph{Original question (null prompts, i.e., $\varnothing$)}
We first query a teacher model \(T\) (DeepSeek\mbox{-}R1~\citep{guo2025deepseek}) on the \emph{unmodified} question text, which we view as a null principle \(p=\varnothing\) with
\begin{equation}
q_i^{(\varnothing)} = q_i.
\end{equation}
where \(q_i^{(\varnothing)}\) is its null-prompt form (unchanged).
For each \(q_i^{(\varnothing)}\) the teacher returns a reasoning trace \(r_i^{(\varnothing)}\) and a full response \(y_i^{(\varnothing)}\) yielding
\begin{equation}
\mathcal{D}_{\text{seed}}
=\Big\{ \big(q_i^{(\varnothing)},\, y_i^{(\varnothing)},\, r_i^{(\varnothing)},\, a_i\big) \Big\}_{i=1}^{N}.
\end{equation}
where 
\(r_i^{(\varnothing)}\) is the teacher-generated reasoning trace, 
\(y_i^{(\varnothing)}\) is the full teacher response, 
\(a_i\) is the ground-truth answer, and 
\(N=90\) is the total number of seed problems.
\paragraph{Selected/Core principle transformations} As shown in Sec.~\ref{selection} and Table~\ref{tab:principles_overview}, we select four principles as our {\em core set}. 
Each principle \(p\in\mathcal{P}_{\text{core}}\) is implemented as a deterministic operator \(f_p(\cdot)\) that \emph{wraps} the original text with a fixed instructional template \(\tau_p\) while leaving the tokens of \(q_i\) unmodified:
\begin{equation}
q_i^{(p)} = f_p(q_i) = \tau_p \,\Vert\, q_i \quad \text{(string concatenation \(\Vert\))}.
\end{equation}
where \(q_i^{(p)}\) is the wrapped question under principle \(p\), and \(\tau_p\) is the fixed instructional template. This construction preserves the mathematical content, since removing \(\tau_p\) recovers the original question \(q_i\). Querying \(T\) with each \(q_i^{(p)}\) produces \((r_i^{(p)}, y_i^{(p)})\), and we collect
\begin{equation}
\mathcal{D}_{\text{core}}=
\Big\{\big(q_i^{(p)},\, r_i^{(p)},\,  y_i^{(p)},\, a_i\big)\; \Big|\; i=1,\dots,N;\; p\in\mathcal{P}_{\text{core}}\Big\}.
\end{equation}
where 
\(r_i^{(p)}\) is the teacher reasoning trace, 
\(y_i^{(p)}\) is the teacher response, 
\(a_i\) is the ground-truth answer, 
\(N\) is the number of seed problems, 
and \(\mathcal{P}_{\text{core}}\) is the set of four selected principles.

\paragraph{Reward framing variants for {\em reward principle}} Our single‑principle ablation (Sec.~\ref{sec:ablations}) shows that \emph{Reward Framing} yields the largest accuracy gain among the core strategies. To test whether this effect depends on exact wording, we create six paraphrased Reward prompts that vary in incentive magnitude, placement, and phrasing strength.\footnote{Appendix~\ref{appendix:p6_variants} reports model accuracy across five Reward Framing variants (R1--R5) that differ in reward amount (e.g., \$20 vs. \$200,000) and location (e.g., beginning vs. end of prompt). Results are presented for o1-mini, Gemini, and DeepSeek.}
$
\mathcal{V}_{\text{Reward}} = \{R_1, R_2, R_3, R_4, R_5, R_6\}.
$
Applying each Reward variant to all \(N\) problems and querying \(T\) gives :
\begin{equation}
    \mathcal{D}_{\text{Reward}} =
\Big\{ \big(\hat{q}_i^{(R_j)}, r_i^{(R_j)},  y_i^{(R_j)}, a_i^\ast \big)
\;\Big|\; i = 1,\dots,N;\; R_j \in \mathcal{V}_{\text{Reward}} \Big\}.
\end{equation}
where \(\hat{q}_i^{(R_j)}\) is the problem \(q_i\) wrapped with Reward variant \(R_j\) and 
\(\mathcal{V}_{\text{Reward}}\) the set of six Reward Framing variants.
\noindent
Table~\ref{tab:principles_overview} summarizes the four core principles, and Table~\ref{tab:reward_variants} lists the Reward Framing paraphrases in full. 

\noindent{\bf Prompt order impact.} We also study how the ordering of prompts affects LLM responses and, in turn, their reasoning ability. Specifically, we experiment with placing the additional prompt either before (R1) or after (R4) the original question, as shown in Table~\ref{tab:reward_variants}. In general, positioning prompts at the beginning leads the model to focus more on the prompt's instructions and therefore achieves better accuracy, as shown in~Table~\ref{tab:p6_variations_accuracy}. Intriguingly, we further notice that when the question is relatively short, the difference in accuracy is actually small.

\begin{table}[ht]
\centering
\small
\setlength{\tabcolsep}{6pt}
\begin{tabular}{clc}
\toprule
\textbf{Variant ID} & \textbf{Incentive Phrase} & \textbf{Transformation} \\
\midrule
\textbf{R1} & I am going to tip \$200{,}000 for a better solution! & $\mathcal{I}(R_1)$ \\
\textbf{R2} & I am going to tip you \$200{,}000 \textbf{if you provide the correct answer!} & $\mathcal{T}_{\mathrm{COND}}(R_1)$ \\
\textbf{R3} & I am going to tip you \textbf{\$200 if you provide a better solution!} & $\mathcal{T}_{\mathrm{COND}}(\mathcal{T}_{\mathrm{LOW}}(R_1))$  \\
\textbf{R4} & \textbf{Add at the end of the prompt}: I am going to tip \$200{,}000 for a better solution!  & $\mathcal{T}_{\mathrm{SUFFIX}}(R_1)$ \\
\textbf{R5} &  I am going to tip \textbf{\$200} for a better solution! & $\mathcal{T}_{\mathrm{LOW}}(R_1)$ \\
\textbf{R6} &  I am going to tip you \$500{,}000 for a better solution and \textbf{step-by-step reasoning!} & 
$\mathcal{T}_{\mathrm{REASON}}(\mathcal{T}_{\mathrm{HIGH}}(R_1))$
 \\
\bottomrule
\end{tabular}
\caption{
\textbf{
Reward Framing} variants in \textbf{\texttt{P-TTS}}. 
Each variant is generated from the baseline $R_1$ via an operator: $R_j=\mathcal{T}_j(R_1)$, where $\mathcal{I}$ denotes identity. 
Operator definitions: 
$\mathcal{T}_{\mathrm{COND}}\!: C$ (\emph{add conditionality});\;
$\mathcal{T}_{\mathrm{LOW}}\!: M{\downarrow}$ (\emph{reduce magnitude in USD});\;
$\mathcal{T}_{\mathrm{HIGH}}\!: M{\uparrow}$ (\emph{increase magnitude in USD});\;
$\mathcal{T}_{\mathrm{SUFFIX}}\!: \pi{=}\text{suffix}$ (\emph{change placement});\;
$\mathcal{T}_{\mathrm{REASON}}\!: \rho{\neq}\varnothing$ (\emph{add reasoning cue, e.g., step-by-step}).
}

\label{tab:reward_variants}
\end{table}

\paragraph{Dataset configurations and scale}
We construct \emph{four} dataset families from the same $N$ seeds, distinguishing single–principle sets from their unions:

\begin{enumerate}[label=(\roman*), leftmargin=2.2em]
  \item \textbf{Single-P\textnormal{-}TTS (per principle).}
 For each $p\in\mathcal{P}_{\text{core}}$ we build a \emph{separate} dataset
\[
\mathcal{D}^{\text{single}}_{(p)} \;=\; \big\{\,\big(q_i^{(p)},\, y_i^{(p)},\, r_i^{(p)},\, a_i^\ast\big)\,\big\}_{i=1}^{N},
\quad\big|\mathcal{D}^{\text{single}}_{(p)}\big|=N=90.
\]
Thus, there are four Single-\texttt{P-TTS} datasets (one per principle). When we report “Single”, we train one model per $p$ (and report results per $p$ or their mean, as specified in Sec.~\ref{sec:ablations}).
  \item \textbf{Core-\texttt{P-TTS} (union of singles).}
  The core set is the (disjoint) union over all four principles:
\[
\mathcal{D}_{\text{core}} \;=\; \bigsqcup_{p\in\mathcal{P}_{\text{core}}}\mathcal{D}^{\text{single}}_{(p)},
\qquad |\mathcal{D}_{\text{core}}| = 4N = 360.
\]

  \item \textbf{Seed combined with the core \texttt{P-TTS}.}
  We add the null-prompt (seed)  $\mathcal{D}_{\text{seed}}$ and $\mathcal{D}_{\text{Reward}}$ to obtain
\[
\mathcal{D}_{\text{seed+core}} \;=\; \mathcal{D}_{\text{seed}} \cup \mathcal{D}_{\text{core}},
\qquad |\mathcal{D}_{\text{seed+core}}| = 5N = 450.
\]

  \item \textbf{Full \texttt{P-TTS}.}
Let $\mathcal{V}_{\text{Reward}}$ denote the set of reward paraphrases $K$ total variants applied to all seeds, producing $\mathcal{D}_{\text{Reward}}$. Because one variant is already used in the Core set, the additional Reward set has size $(K-1)N$.
The full corpus is
\[
\mathcal{D}_{\text{full\;P\textnormal{-}TTS}} \;=\; \mathcal{D}_{\text{seed}} \cup \mathcal{D}_{\text{core}} \cup \mathcal{D}_{\text{Reward}},
\qquad |\mathcal{D}_{\text{full\;P\textnormal{-}TTS}}| = (1+4+(K-1))N.
\]
In our experiments we use $K{=}6$, so 
$|\mathcal{D}_{\text{full\;P\textnormal{-}TTS}}| = 10N = 900.
$
\end{enumerate}

\noindent
We parameterize the corpus size by the augmentation multiplier \(m \coloneqq |\mathcal{D}|/N\), where \(\mathcal{D}\) is the training corpus. In our study \(|\mathcal{D}|\in\{90,360,450,900\}\) with \(N{=}90\), so \(m\in\{1,4,5,10\}\), corresponding to Single, Core, Seed+Core, and Full, respectively, enabling controlled comparisons as a function of prompt diversity (Fig.~\ref{fig:agum-multi}).

\subsection{Fine-Tuning with \texttt{P-TTS} Dataset Augmentations}

We evaluate whether principle-guided wrapping improves supervised reasoning under constrained data. We fine-tune \texttt{Qwen2.5-Instruct} (7B/14B/32B) \citep{qwen25} separately on each configuration from Sec.~\ref{sec:ptts_data_construction}, following an SFT recipe adapted from s1 \citep{muennighoff2025s1}. The student is trained to predict full assistant outputs (reasoning + answer) with token-level cross-entropy computed on assistant tokens only (user tokens masked). Each dataset configuration (Original, Single, Core, Mix, and  Full \texttt{P-TTS}) is used to train a separate model, enabling us to isolate the contribution of each prompting strategy and the effect of data scaling, i.e., isolating the impact of instructional diversity and corpus scale ($m\in\{1,4,5,10\}$) while keeping optimization and decoding fixed across runs. 

Models are trained to predict the full assistant output—reasoning trace and final answer.
Our dataset scale (\(90 \rightarrow 900\)  examples) is intentionally small, allowing us to directly measure how principle‑guided prompt reformulations affect supervised reasoning performance relative to models trained on much larger datasets.

\begin{algorithm}[t]
\caption{Prompting Test-Time Scaling (\texttt{P-TTS}) Dataset Construction}
\label{alg:ptts_core}
\DontPrintSemicolon
\KwIn{Seeds $\mathcal{O}_{\mathrm{seed}}=\{(q_i,a_i\}_{i=1}^N$; core principles $\mathcal{P}_{\mathrm{core}}$; reward variants $\mathcal{V}_{\texttt{Reward}}$; teacher $T$}
\KwOut{Augmented dataset $\mathcal{D}_{\mathrm{\text{full\;P\textnormal{-}TTS}}}$}
$\mathcal{D}_{\mathrm{\text{full\;P\textnormal{-}TTS}}} \leftarrow \emptyset$ \;
\ForEach{$(q_i,a_i) \in \mathcal{O}_{\mathrm{seed}}$}{
  Query $T$ with $q_i$ to obtain $(r_i^{(\varnothing)},y_i^{(\varnothing)})$ \;
  Add $(q_i, r_i^{(\varnothing)}, y_i^{(\varnothing)}, a_i)$ to $\mathcal{D}_{\mathrm{\text{full\;P\textnormal{-}TTS}}}$ \;
  \ForEach{$p \in \mathcal{P}_{\mathrm{core}}$}{
    $q_i^{(p)} \leftarrow \tau_p \,\Vert\, q_i$ \tcp*[r]{wrap; preserve $q_i$}
    Query $T$ with $q_i^{(p)}$ to obtain $(r_i^{(p)},y_i^{(p)})$ \;
    Add $(q_i^{(p)}, r_i^{(p)}, y_i^{(p)}, a_i)$ \;
  }
  \ForEach{$R_j \in \mathcal{V}_{\texttt{Reward}}$}{
    $q_i^{(R_j)} \leftarrow \tau_{R_j} \,\Vert\, q_i$ \;
    Query $T$ with $q_i^{(R_j)}$ to obtain $(r_i^{(R_j)},y_i^{(R_j)})$ \;
    Add $(q_i^{(R_j)}, r_i^{(R_j)}, y_i^{(R_j)}, a_i)$ \;
  }
}
\Return{$\mathcal{D}_{\text{full\;P\textnormal{-}TTS}}$}
\end{algorithm}

\section{Experiments}

\subsection{Experimental Setup}
\label{sec:experimental_setup}

\textbf{Datasets.} We evaluate our \textbf{\texttt{P-\-TTS}} models on four public reasoning benchmarks: \textbf{AIME24}~\citep{2024AIME} (30 problems) and \textbf{AIME25}~\citep{2025AIME} (15 problems) from the American Invitational Mathematics Examination; AIME includes problems from algebra, arithmetic, geometry, number theory, combinatorics, and probability. \textbf{MATH500}~\citep{hendrycks2021measuring} is a 500‑problem competition‑math subset; we adopt the publicly released OpenAI selection used in prior work. \textbf{GPQA-\-Diamond}~\citep{rein2024gpqa} contains 198 PhD‑level science questions from Biology, Chemistry, and Physics with reported expert performance of 69.7\%. We evaluate using the \texttt{lm-evaluation-harness} framework~\citep{gao2021lmevalharness,biderman2024lessons}. To make results comparable across models and ablations, we disable sampling by setting the temperature to 0 (greedy decoding), so each input yields a deterministic output. Reported scores are accuracy (equivalent to pass@1). In addition to these four core benchmarks, we further assess cross-domain and multilingual generalization using a broader set of reasoning tasks spanning Chinese exams, U.S. school math, olympiad-style problems, and scientific quantitative reasoning. These evaluations, shown in Table~\ref{tab:extra}, include \textbf{Gaokao}, \textbf{Kaoyan}, \textbf{OlympiadBench}~\cite{he2024olympiadbench}, \textbf{AMC23}, \textbf{GradeSchoolMath}, and \textbf{Minerva}. 
\newline
\noindent\textbf{Baselines.} 
We benchmark \textbf{\texttt{P‑TTS}} against three categories of reasoning models. \emph{(i) Closed‑source (API‑only) models}: OpenAI’s o1 series~\citep{openai2024reasoning,openai_o3_mini} and Google’s experimental Gemini 2.0 Flash Thinking variant~\citep{google2024vertexai}.  
\emph{(ii) Open‑weight models}: DeepSeek‑R1 series \citep{guo2025deepseek} and  Qwen’s QwQ‑32B‑preview~\citep{qwq-32b-preview,qwen25} .  
\emph{(iii) Open‑weight SFT models on Qwen2.5‑Instruct with public data on openly available reasoning corpora}: including Bespoke‑Stratos‑32B~\citep{bespoke_stratos_32b}, OpenThinker‑32B~\citep{openthinker32b_hf,guha2025openthoughts}, Sky‑T1‑32B‑Preview~\citep{sky_t1_2025}, and the S1/ S1.1 w/o BF \citep{muennighoff2025s1}checkpoints.  
\newline
\noindent\textbf{Diversity Metrics.} We compute two complementary metrics—\emph{semantic} and \emph{surface–level}—to quantify how each single-principle variant in \(\mathcal{D}_{\text{core}}\) adds information beyond the seed set (\(\mathcal{D}_{\text{seed}}\)). \textbf{Semantic diversity (Diversity Gain).} Following \citep{yu2023metamath,bilmes2022submodularity}, we compute \emph{diversity gain} to quantify knowledge-level novelty. Given a seed dataset \(\mathcal{D}_{\text{seed}}\) and a new dataset \(\mathcal{D}_{\text{core}}\), we define \(\text{DG} = \frac{1}{M} \sum_{x_i \in \mathcal{D}_{\text{core}}} \min_{x_j \in \mathcal{D}_{\text{seed}}} \big\|f(x_i) - f(x_j)\big\|_2^2\), where \(f(\cdot)\) is an embedding function and \(M = \lvert \mathcal{D}_{\text{core}} \rvert\). We use OpenAI’s \texttt{text-embedding-ada-002} as \(f\) for feature extraction. Higher values indicate greater semantic divergence from the base data. \textbf{Surface–level diversity (trigram diversity).} We compute \emph{trigram diversity}~\cite{li-etal-2022-evade}, defined as the ratio of non-overlapping trigrams between two texts\footnote{For texts $x,y$, $\mathrm{TD}(x,y)=1-\frac{|\Tri(x)\cap \Tri(y)|}{|\Tri(x)\cup \Tri(y)|}$, where $\Tri(x)$ denotes the set of \emph{distinct} word-level trigrams in $x$.}.
 We average this score over all sample pairs between each \texttt{P–TTS} principle variant in \(\mathcal{D}_{\text{core}}\) and its corresponding baseline instance in \(\mathcal{D}_{\text{seed}}\).

\begin{figure}[ht]
  \centering
  \includegraphics[width=0.8\linewidth]{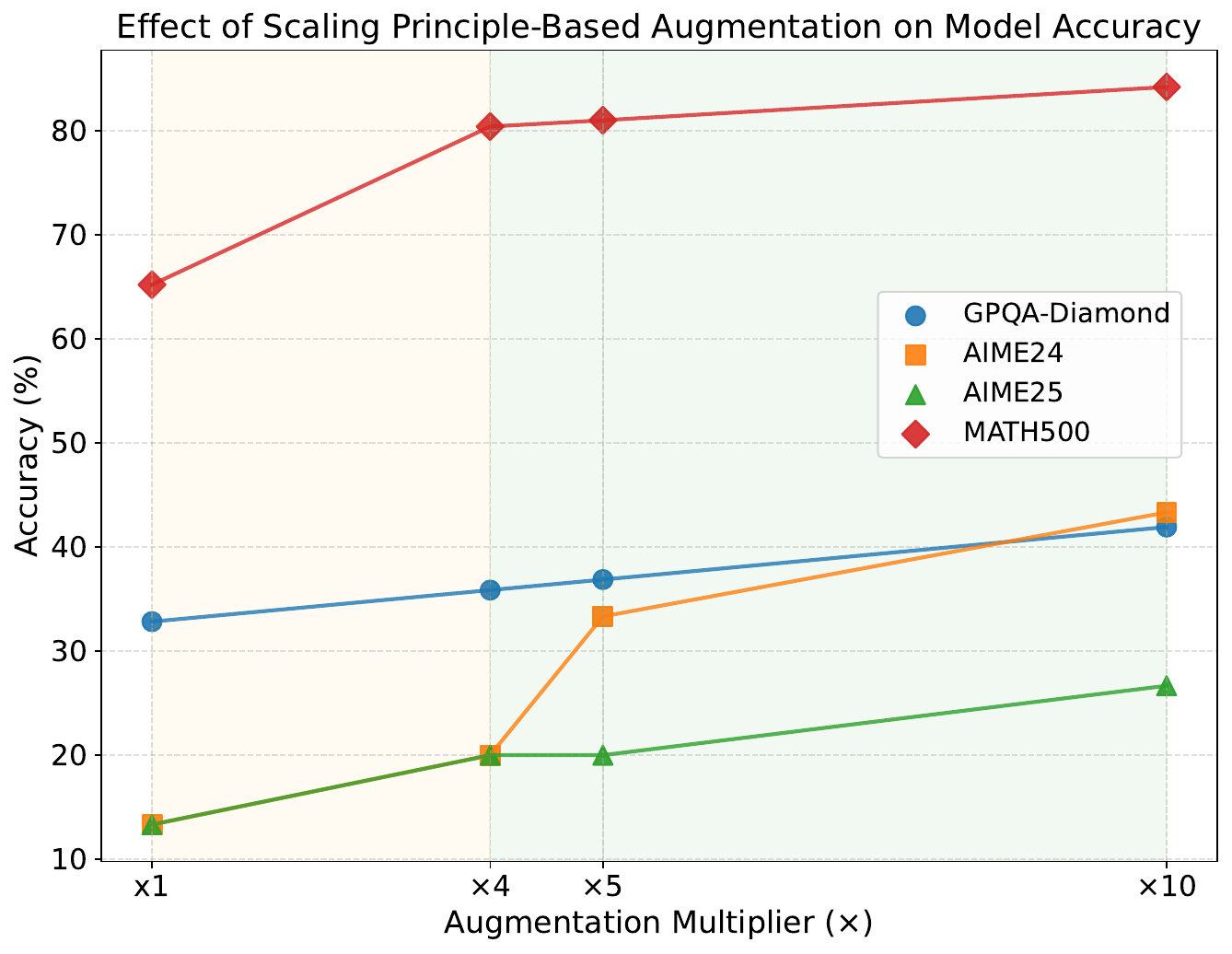}
    \caption{\textbf{Accuracy improvement with increased principled augmentation on 7B model.}  
  We evaluate how model accuracy scales with the number of augmented training examples. 
  Here, \textbf{$\times$1} refers to \texttt{P-TTS\textsubscript{Reward}} (90 examples), 
  \textbf{$\times$4} to \texttt{P-TTS\textsubscript{Core}} (360 examples), 
  \textbf{$\times$5} to \texttt{P-TTS\textsubscript{Core+Orig}} (450 examples), 
  and \textbf{$\times$10} to \texttt{P-TTS\textsubscript{Full}} (900 examples). 
  Accuracy improves consistently across all evaluation sets with larger, principle-guided augmentations.}
  \label{fig:agum-multi}
\end{figure}
\subsection{Teacher Model for Data Construction}
Our objective is to construct a \textbf{compact yet high-quality} mathematical reasoning corpus for supervised fine-tuning.  Specifically, we aim to identify the possible \textbf{smallest}  training corpus that still yields the highest downstream accuracy. To obtain both full responses and explicit reasoning traces, we consider large language models that expose chain-of-thought generation through their public APIs. We benchmark three reasoning models: Claude-3-Opus, DeepSeek-R1, and OpenAI Omni-4 on four tasks. Claude-3-Opus and DeepSeek-R1 natively return aligned answer–reasoning pairs, while Omni-4 requires an augmented prompt to elicit full reasoning.  
From each model, we collect 90 answer–reasoning pairs and fine-tuned a Qwen2.5-7B-Instruct on the resulting dataset.  Table~\ref{tab:model-comparison} reports accuracy averaged over all four benchmarks.  
The Qwen2.5-7B-Instruct model fine-tuned on DeepSeek-R1 outputs consistently outperforms counterparts trained on Claude and Omni-4 outputs under the same small corpus.
Based on this, we adopt \textbf{DeepSeek-R1} as the teacher for all subsequent data construction.

\begin{figure}[ht]
    \centering
    \setlength{\tabcolsep}{2pt}  
    \renewcommand{\arraystretch}{1}  
    \begin{tabular}{cc}
        \begin{subfigure}[b]{0.48\textwidth}
            \includegraphics[width=\linewidth]{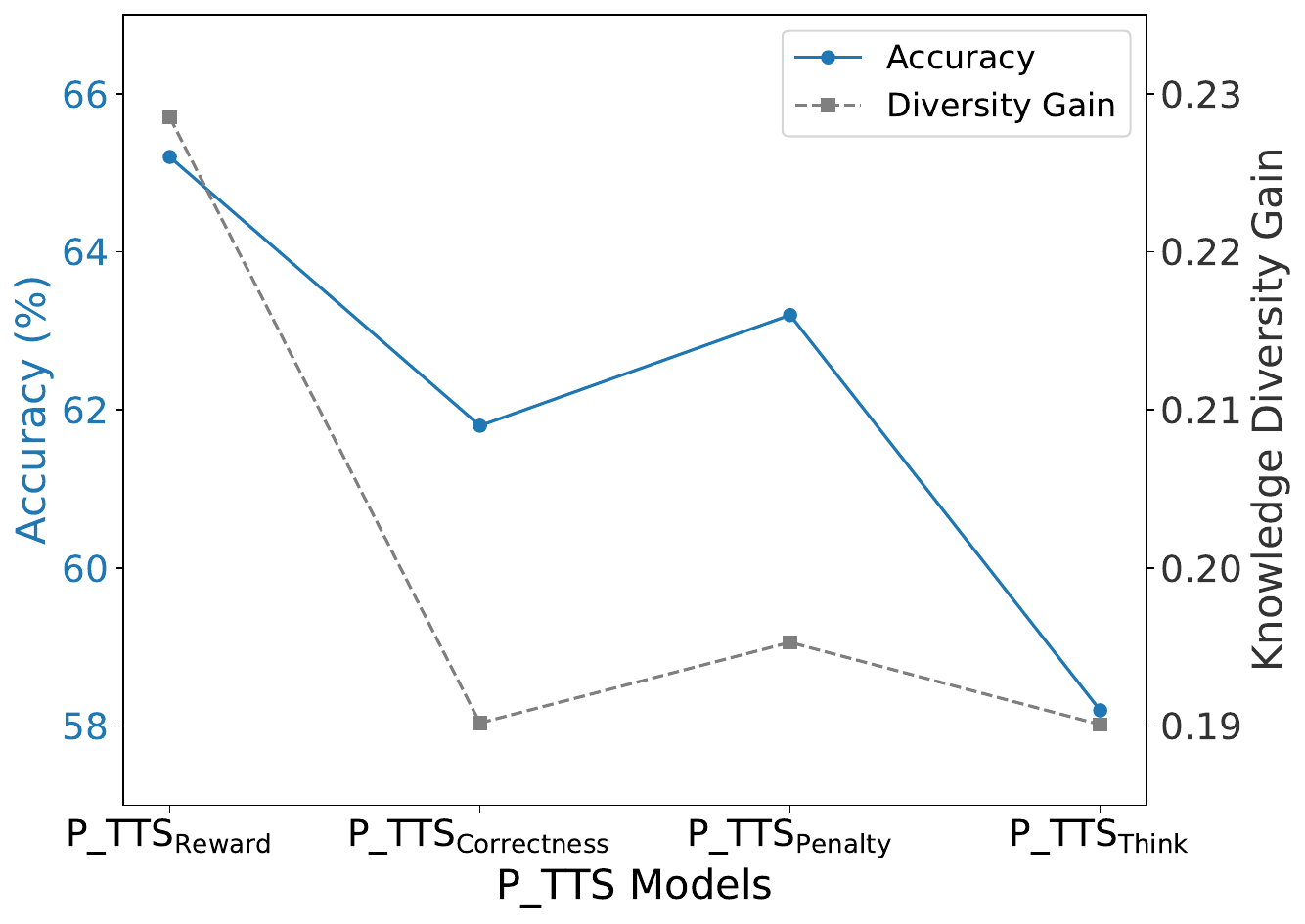}
            \caption{MATH500}
            \label{fig:dataset1}
        \end{subfigure} &
        \begin{subfigure}[b]{0.48\textwidth}
            \includegraphics[width=\linewidth]{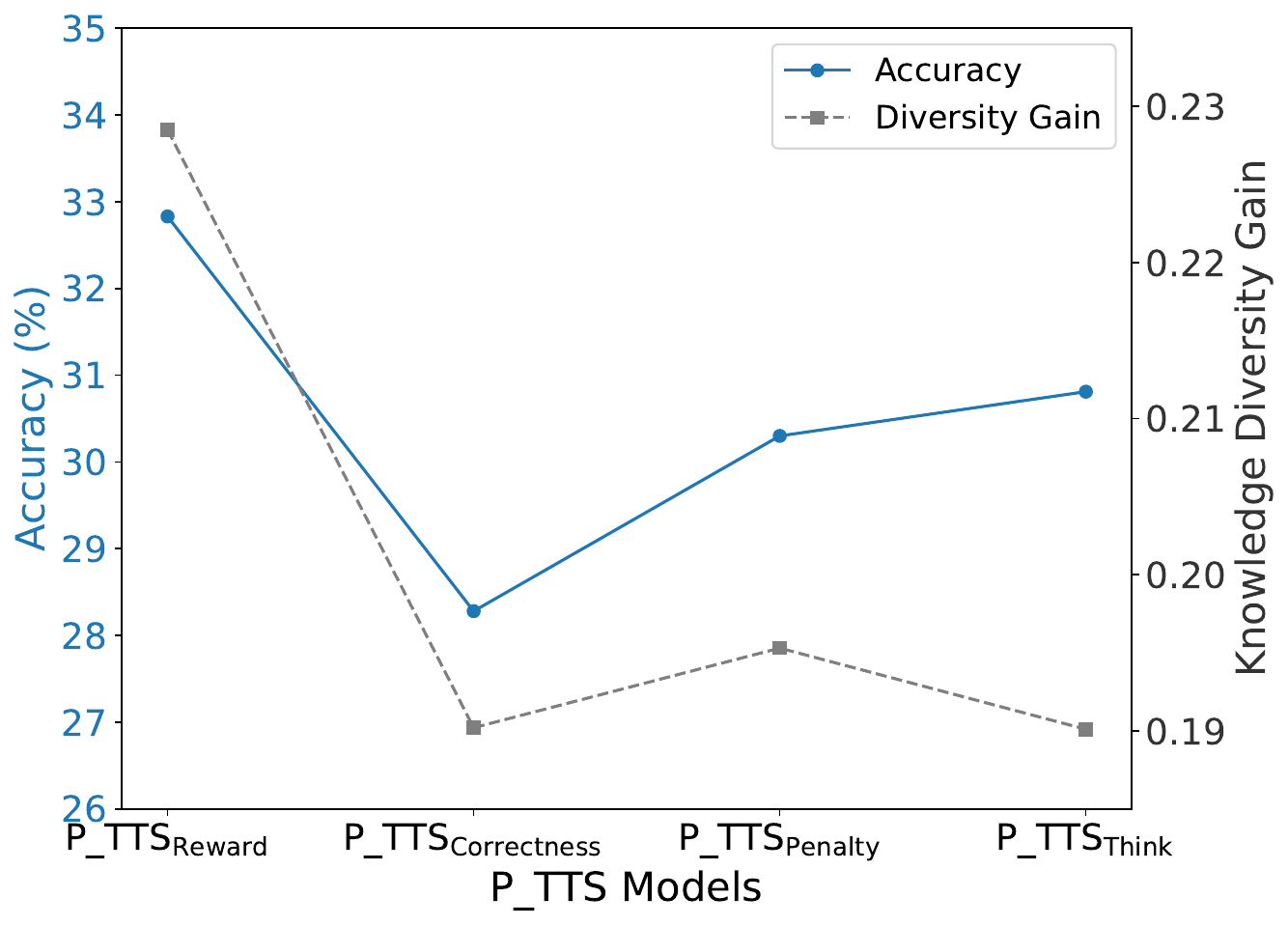}
            \caption{GPQA-Diamond}
            \label{fig:dataset2}
        \end{subfigure} \\
        \begin{subfigure}[b]{0.48\textwidth}
            \includegraphics[width=\linewidth]{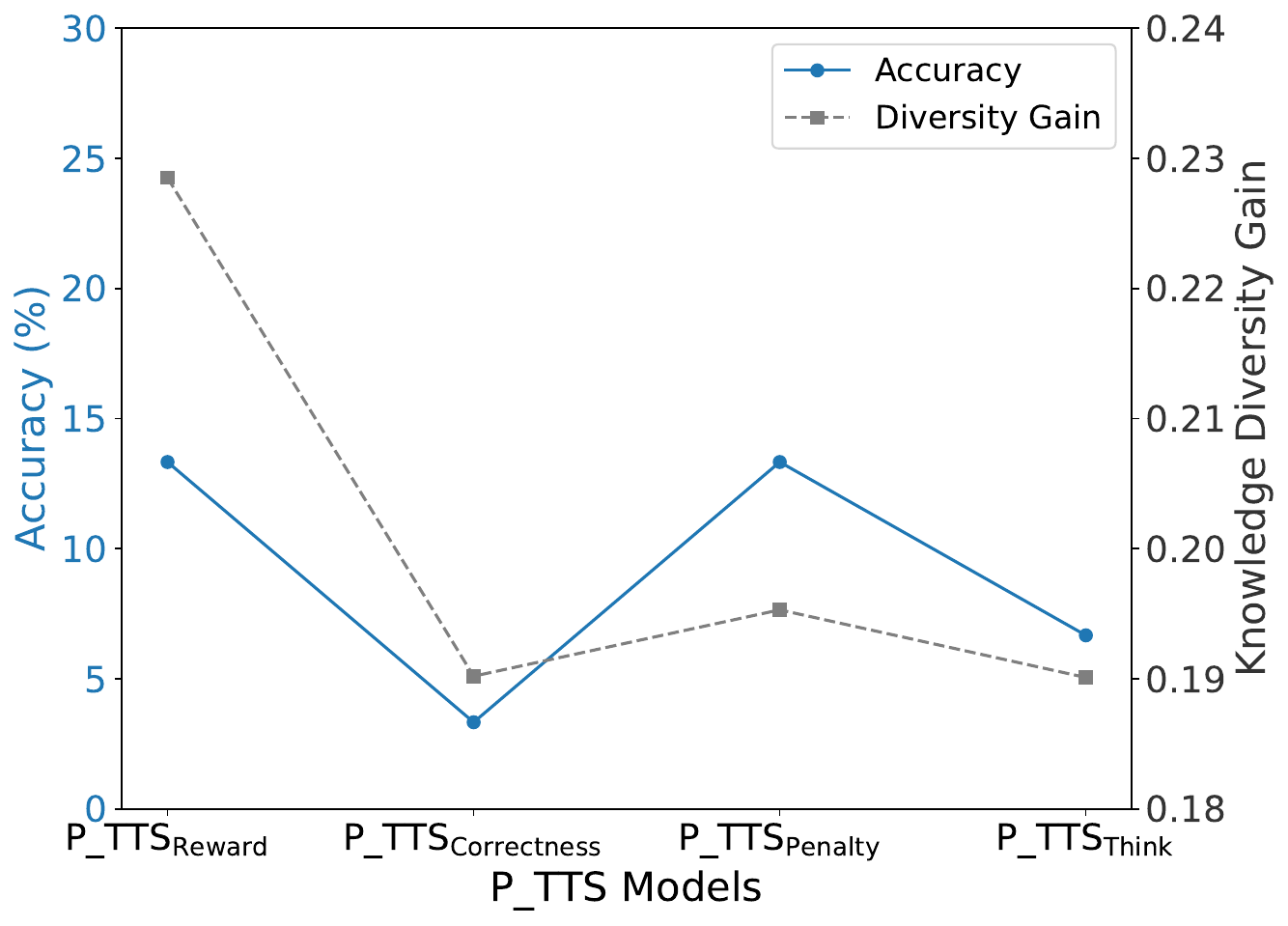}
            \caption{AIME24}
            \label{fig:dataset3}
        \end{subfigure} &
        \begin{subfigure}[b]{0.48\textwidth}
            \includegraphics[width=\linewidth]{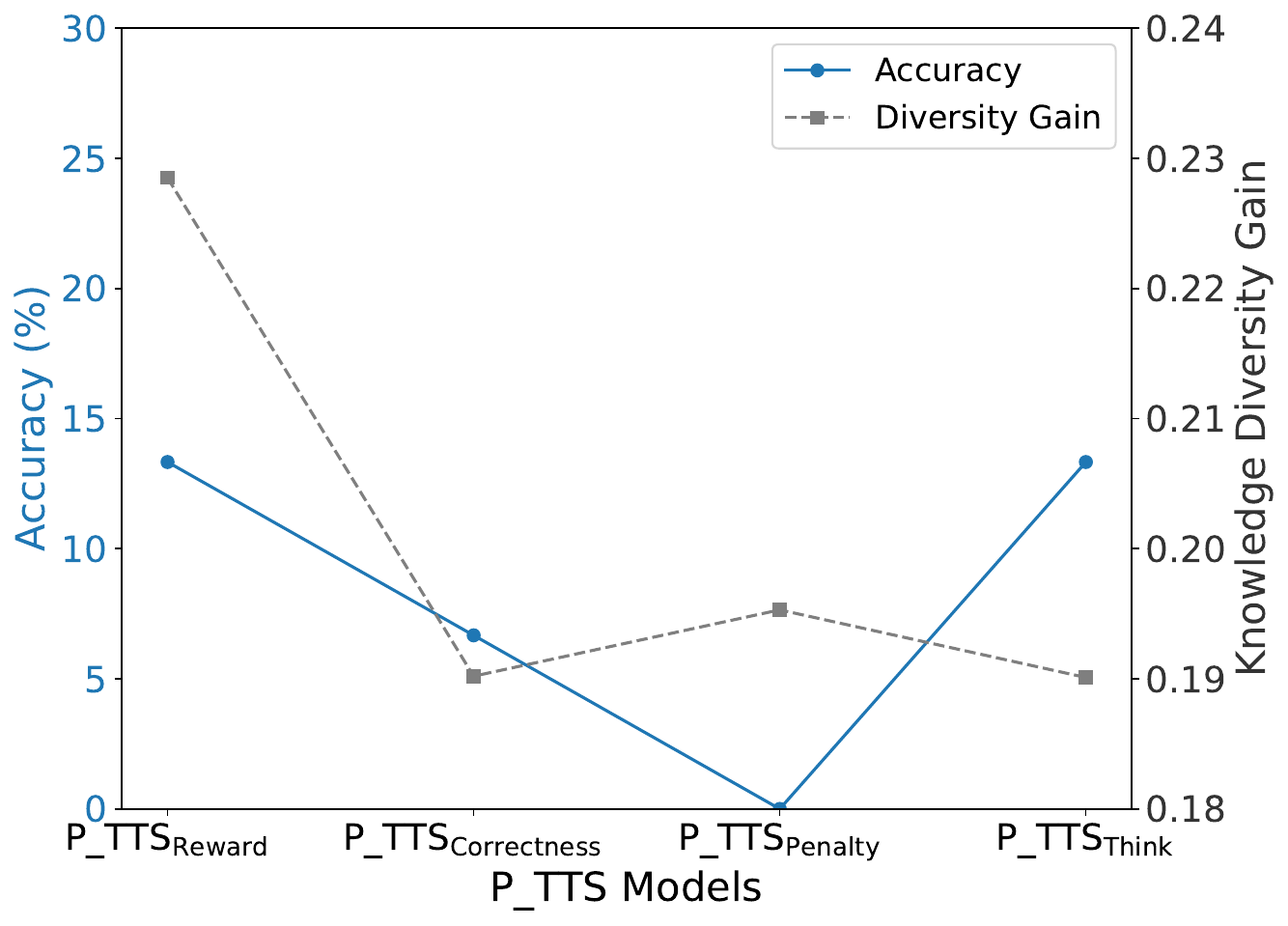}
            \caption{AIME25}
            \label{fig:dataset4}
        \end{subfigure}
    \end{tabular}
    \caption{\textbf{Knowledge Diversity Gain vs. Accuracy for different \texttt{P-TTS} variants across four benchmarks.} We compare the trade-off between Accuracy (blue solid line, left y-axis) and Knowledge Diversity Gain (gray dashed line, right y-axis) on 7B model for four principled prompting strategies: \texttt{Reward}, \texttt{Correctness}, \texttt{Penalty}, and \texttt{Think}. Diversity Gain is computed relative to the original \texttt{P-TTS} baseline.}

    \label{fig:all_datasets_diversity}
\end{figure}

\subsection{Ablation Studies}
\label{sec:ablations}
\subsubsection{Single‑\texttt{P-TTS}: Measuring the Impact of Each Principle}

\paragraph {Single‑\texttt{P-TTS}}To assess the impact of each principle independently, we fine-tune separate  \texttt{Qwen2.5-7B-Instruct} models on the \textbf{Single \texttt{P-TTS}} subsets—90 examples per model—each applying only one core principle \(p \in \mathcal{P}_{\text{core}}\). We compare these models to a baseline trained on the same seed problem without any instructional framing (\texttt{P-TTS\textsubscript{Seed}}).
As shown in Table~\ref{tab:ptts_single_pairwise}, all \textbf{Single \texttt{P-TTS}} variants outperform the baseline on average. The \texttt{Reward}-based model (\texttt{P-TTS\textsubscript{Reward}}) yields the highest overall gain (+6.67\%), improving accuracy across all benchmarks. \texttt{Penalty}-based model (\texttt{P-TTS\textsubscript{Penalty}}) also delivers strong results, especially on \textsc{MATH500}, though with a drop on \textsc{AIME25}. \texttt{Correctness}-based model (\texttt{P-TTS\textsubscript{Correctness}}) offers modest improvements, while \texttt{Think}-based model (\texttt{P-TTS\textsubscript{Think}}) increases the average but underperforms on \textsc{MATH500}.
These findings highlight that even minimal augmentations, i.e., just 90 principle-guided examples, can yield measurable improvements. Among the four principles, \texttt{Reward} and \texttt{Penalty} framings are the most effective when applied independently.

\begin{table}[ht]
\centering
\begin{adjustbox}{max width=\linewidth}
\begin{tabular}{lcccccc}
\toprule
\textbf{Model} & \textbf{\#Ex.} & \textbf{AIME‑24} & \textbf{AIME‑25} & \textbf{MATH500} & \textbf{GPQA‑D} & \textbf{Avg.} \\
\midrule
\multicolumn{7}{c}{\textbf{Single-Principle \texttt{P-TTS} Ablation (90 Examples Each)}} \\
\midrule
\texttt{P-TTS\textsubscript{Seed}} (baseline)       & 90  &  3.33 &  6.67 & 60.20 & 27.78 & 24.50 \\
\texttt{P-TTS\textsubscript{Reward}}                & 90  & \textbf{13.33} & \textbf{13.33} & \textbf{65.20} & \textbf{32.83} & \textbf{31.17} \\
\texttt{P-TTS\textsubscript{Correctness}}           & 90  &  3.33 &  6.67 & 61.80 & 28.28 & 25.02 \\
\texttt{P-TTS\textsubscript{Penalty}}               & 90  & 13.33 &  0.00 & 63.20 & 30.30 & 26.71 \\
\texttt{P-TTS\textsubscript{Think}}                 & 90  &  6.67 & \textbf{13.33} & 58.20 & 30.81 & 27.25 \\
\midrule
\multicolumn{7}{c}{\textbf{Pairwise Principle Ablation Centered on Reward Framing (180 Examples Each)}} \\
\midrule
\texttt{P-TTS}\(_{\text{Reward}\cup\text{Penalty}}\)          & 180 & 10.00 & \textbf{20.00} & 75.20 & 32.32 & 34.38 \\
\texttt{P-TTS}\(_{\text{Reward}\cup\text{Correctness}}\)      & 180 & \textbf{23.33} & \textbf{20.00} & \textbf{75.40} & \textbf{37.37} & \textbf{39.02} \\
\texttt{P-TTS}\(_{\text{Reward}\cup\text{Think}}\)            & 180 & 13.33 & 13.33 & 72.80 & 31.82 & 32.82 \\
\bottomrule
\end{tabular}
\end{adjustbox}
\caption{\textbf{Accuracy (\%) of Single and Pairwise Principle Ablation using 7B model.} 
Top: Single-principle \texttt{P-TTS} variants each trained on 90 instructional prompts.  
Bottom: Pairwise ablations centered on Reward framing, trained on 180 prompts combining Reward with one other principle.  
All models use \texttt{Qwen2.5‑7B‑Instruct} fine-tuning.}
\label{tab:ptts_single_pairwise}
\end{table}

\paragraph {Pairwise‑\texttt{P-TTS}}To further evaluate the usefulness of each principle and whether they can be effectively combined, we explore the case where a principle that is not highly effective on its own might still contribute positively when paired with another. Specifically, since \texttt{P-TTS\textsubscript{Reward}}  shows the strongest performance in the \textbf{Single \texttt{P-TTS}} ablation, we fix the \texttt{Reward} framing and incrementally add one additional principle at a time, resulting in datasets of 180 examples. Table~\ref{tab:ptts_single_pairwise} shows the performance of each pairwise combination. We observe that combining \texttt{Reward} framing with either \texttt{Correctness} or \texttt{Penalty} significantly boosts performance across most benchmarks, particularly on MATH500 and AIME2024. Also, the \texttt{P-TTS}\(_{\text{Reward}\cup\text{Correctness}}\) combination achieves the highest overall accuracy (39.02\%), suggesting a synergistic effect between reward framing and correctness emphasis. In contrast, combining \texttt{Reward} framing with \texttt{Thinking} yields only modest improvements, indicating diminishing returns when both principles primarily influence the reasoning process.

\begin{figure*}[!htb]
  \centering
  \includegraphics[width=0.8\linewidth]{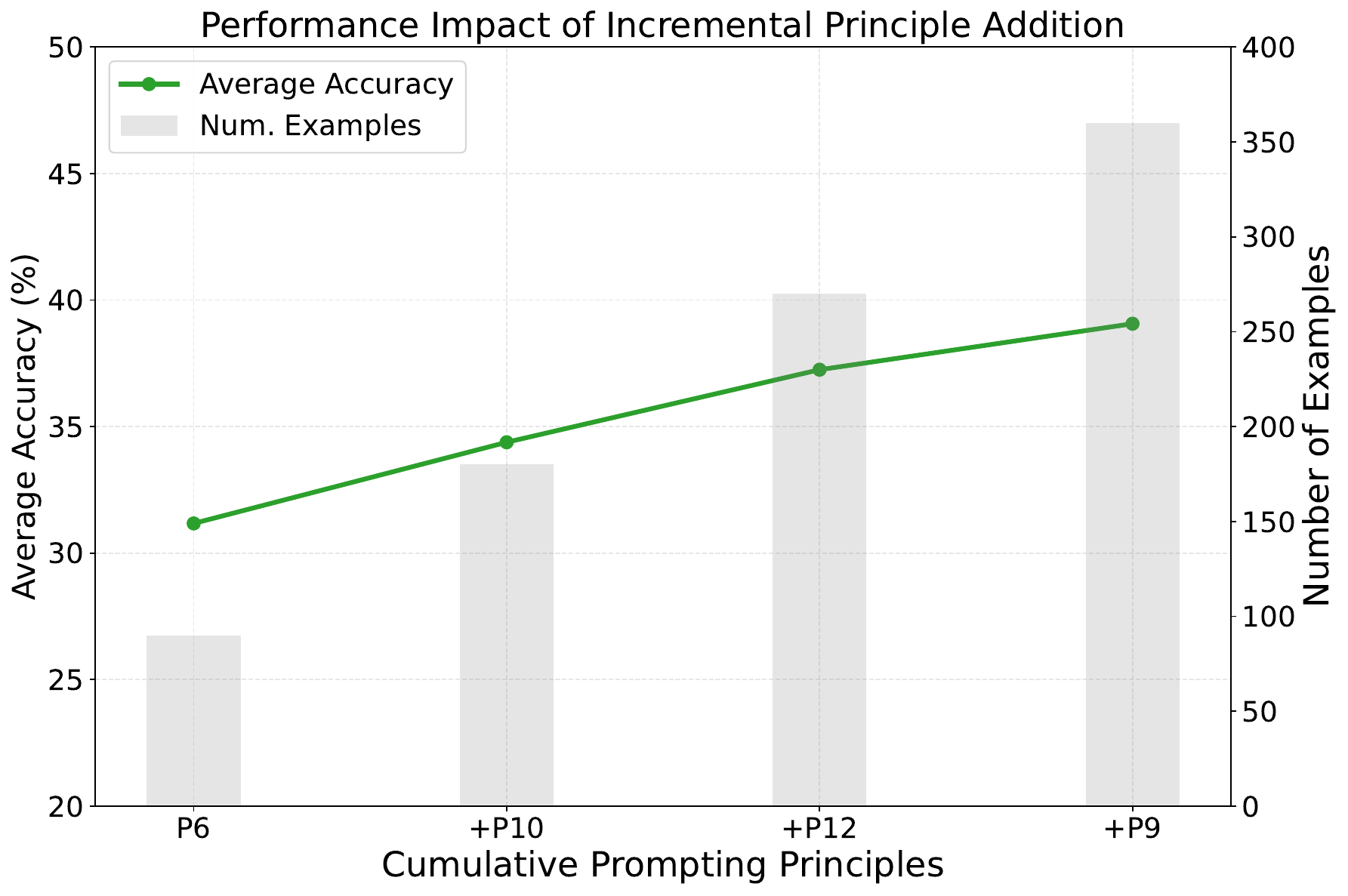}
  \caption{\textbf{Impact of incremental principle addition on average accuracy.}  
  As additional prompting principles are cumulatively incorporated into training (\textbf{P6} → \textbf{+P10} → \textbf{+P12} → \textbf{+P9}), both the number of training examples and model accuracy increase. 
  Bars (right axis) denote the total number of examples after each addition; the green line (left axis) shows the resulting average accuracy across evaluation benchmarks.  
  This highlights the compounding benefit of principled augmentation.}
  \label{fig:abbb}
\end{figure*}

\paragraph {Core‑\texttt{P-TTS}}To assess the relative importance of each principle when used in combination, we conduct a leave-one-out ablation over the full \textbf{Core \texttt{P-TTS}} dataset \(\mathcal{D}_{\text{core}}\) (360 examples). We fine-tune a separate \texttt{Qwen2.5-7B-Instruct} model after removing one principle at a time, reducing the training set to 270 examples. As shown in Table~\ref{tab:leave_one_out}, the exclusion of the \texttt{Reward} framing leads to the largest performance drop (from 39.06\% to 35.40\%), confirming its central role in driving improvements. Removing \texttt{Correctness} or \texttt{Penalty} framing causes moderate degradation, while the absence of \texttt{Step-by-Step thinking} has minimal or slightly positive effects. These results indicate that \texttt{Reward}-based cues are the most impactful when principles are used in combination, whereas \texttt{Step-by-Step} prompting contributes the least in multi-principle settings. Fig.~\ref{fig:abbb} further supports this finding, showing that the incremental addition of principles leads to consistent accuracy improvements.

\begin{table}[ht]
  \centering
  \begin{tabular}{lcccccc}
    \toprule
    \textbf{Model} & \textbf{\#Ex.} & \textbf{AIME2024} & \textbf{AIME2025} & \textbf{MATH500} & \textbf{GPQA‑D} & \textbf{Avg.} \\
    \midrule
    \texttt{P-TTS}\(_{\text{Core}}\) \,(all 4)                      & 360 & 20.00 & 20.00 & 80.40 & 35.86 & 39.06 \\[2pt]
    \midrule
    \texttt{P-TTS}\(_{\text{Core}\setminus\text{Reward}}\)          & 270 & 16.67 & 13.33 & 78.80 & 32.83 & 35.40 \\
    \texttt{P-TTS}\(_{\text{Core}\setminus\text{Correctness}}\)     & 270 & 13.33 & 20.00 & 79.80 & 35.86 & 37.24 \\
    \texttt{P-TTS}\(_{\text{Core}\setminus\text{Penalty}}\)         & 270 & 20.00 & 20.00 & 79.80 & 34.34 & 38.53 \\
   \texttt{P-TTS}\(_{\text{Core}\setminus\text{Think}}\)           & 270 & 20.00 & 26.67 & 78.20 & 32.32 & 39.29 \\
    \bottomrule
  \end{tabular}
    \caption{\textbf{Leave‑one‑principle‑out ablation}.  
           We fine‑tune \texttt{Qwen2.5‑7B‑Instruct} on the full Core‑\texttt{P‑TTS} set  
           (\texttt{Reward+Correctness+Penalty+Think}; 360 prompts) and then re‑train
           after omitting one principle (270 prompts).}
  \label{tab:leave_one_out}
\end{table}

\subsubsection{Measuring the Effects of Augmentation Size}
Table~\ref{tab:data_volume_ablation} presents the results for three configurations: 
(1) \texttt{P-TTS\(_{\text{Core}}\)} with 360 examples, 
(2) \texttt{P-TTS\(_{\text{Core+Seed}}\)} with 450 examples, and (3) \texttt{P-TTS\(_{\text{Core+Seed+RewardVar}}\)} with 900 examples, which represents the full dataset (core, seed, and paraphrastic reward variants). Performance improves consistently with dataset size: the model trained on the full 900-example dataset achieves an average accuracy of 49.03\%, outperforming all other configurations and surpassing the 1k-example S1.1 baseline (38.99\%).
The largest average gain occurs between 450 and 900 examples (+6.23\%), with improvements on all benchmarks.
The increase from 360 to 450 examples is particularly notable on AIME24 (+13.33\%), indicating that mixing seed and wrapped questions with the selected core principles is beneficial.
Fig.~\ref{fig:agum-multi} shows the same trend: scaling augmentation from $\times$1 to $\times$10 yields gains across all four test sets, especially on MATH500 and AIME24.
These results demonstrate that principled prompt augmentation scales effectively and enables competitive performance with far fewer training examples than traditional supervised fine-tuning.

\begin{table}[ht]

  \centering
  \begin{tabular}{lcccccc}
    \toprule
    \textbf{Model} & \textbf{\#Ex} & \textbf{AIME24} & \textbf{AIME25} & \textbf{MATH500} & \textbf{GPQA‑D} & \textbf{Avg.} \\
    \midrule
    Qwen2.5‑7B‑Instruct (base)          & –     & 13.33 &  6.67 & 76.40 & 36.36 & 33.19 \\
    S1‑7B      & 1K & 16.67 & 13.33 & 77.20 & 41.41 & 37.15 \\
    S1.1‑7B   & 1K & 13.33 & 20.00 & 81.20 & 41.41 & 38.99 \\
    \midrule
    \texttt{P\_TTS}\(_\text{Core}\)‑7B                        & 360  & 20.00 & 20.00 & 80.40 & 35.86 & 39.07 \\
   \texttt{P\_TTS}\(_\text{Core+ Seed}\)‑7B                   & 450  & \ \ 33.33$^{\uparrow}$ & 20.00 & \ \ 81.00$^{\uparrow}$ & \ \ 36.87$^{\uparrow}$ & \ \ 42.80$^{\uparrow}$ \\
    P‑TTS\(_{\text{Core+ Seed+ RewardVar}}\)‑7B       & 900  & \ \ \textbf{43.33}$^{\uparrow}$ & \ \ \textbf{26.67}$^{\uparrow}$ & \ \ \textbf{84.20}$^{\uparrow}$ & \ \ \textbf{41.92}$^{\uparrow}$ & \ \ \textbf{49.03}$^{\uparrow}$ \\
    \bottomrule
  \end{tabular}
  \caption{\textbf{Data‑volume ablation.} Accuracy (\%) when fine‑tuning  
           (i) on Core \texttt{P‑TTS} only (360 prompts),  
           (ii) Core+ Seed (450), and  
           (iii) Core+ Seed+ six Reward–variant prompts (900).  
           Best values in each column are \textbf{bold}; $\uparrow$ marks a gain over the immediately preceding configuration.}
  \label{tab:data_volume_ablation}
\end{table}

\subsection{\texttt{P-TTS} Dataset Analysis}
\paragraph{Diversity}
The accuracy results in Table~\ref{tab:ptts_single_pairwise} show that each Single \texttt{P-TTS} variant outperforms the seed baseline on most benchmarks and on average. \texttt{P-TTS}\textsubscript{Reward} yields the largest absolute gain, improving performance by approximately 6.7\% on \textsc{AIME25} and 5.0\% on \textsc{GPQA-Diamond}, while \texttt{P-TTS}\textsubscript{Penalty} achieves the strongest improvement on \textsc{MATH500}. Even the weakest variant, \texttt{P-TTS}\textsubscript{Correctness}, matches or surpasses the baseline on three out of four datasets. These accuracy trends correlate with the diversity analysis in Fig.~\ref{fig:all_datasets_diversity}. Single \texttt{P-TTS} \emph{variants} with higher \emph{semantic diversity} (Diversity Gain), such as \texttt{P-TTS}\textsubscript{Reward} and \texttt{P-TTS}\textsubscript{Penalty}, also exhibit larger accuracy improvements. Table~\ref{tab:trigram_diversity} complements this by quantifying \emph{surface-level diversity} (trigram diversity): \texttt{P-TTS}\textsubscript{Reward} attains the highest scores across both final responses and reasoning traces, consistent with the observed performance gains. This pattern is consistent with prior findings that semantically diverse supervision and diversity-promoting augmentations improve generalization and robustness~\citep{bukharin2024data,qu2021coda}.
\begin{figure}[!h]
\begin{tcolorbox}[colback=baseline!30, colframe=principle1!80!black, title=\texttt{P-TTS}-32B]

  \begin{tcolorbox}[colback=baseline, title=Question]
  If the Sun's absolute magnitude is $+5$, find the luminosity of a star of magnitude $0$ in ergs/s. A useful constant: the luminosity of the sun is $3.83 \\times 10^{33}$ ergs/s
  \end{tcolorbox}

  \begin{tcolorbox}[colback=baseline, title=Response]
 Okay, so I need to find the luminosity of a star with magnitude 0, given that the Sun's absolute magnitude is +5 and its luminosity is $3.83 \times 10^{33}$ ergs/s. Hmm, let's start by recalling the relationship between magnitude and luminosity.  First, absolute magnitude ........ given as +5. The star we're interested in has an absolute magnitude of 0. The difference in magnitude between the two stars is 5 - 0 = 5 magnitudes.  In astronomy, the difference in magnitudes relates to the ratio of luminosities. The formula is:  m1 - m2 = -2.5 log10 (L1 / L2)  ........ indeed m1 - m2 = -2.5 log10 (L1 / L2). So if we know the difference in magnitudes, we can find the ratio of luminosities.  In our case, m1 (Sun) is +5, m2 (star) is 0. So m1 - m2 = 5 - 0 = +5.  Plugging into the formula:  $5 = -2.5 \log_{10} (L_{\text{sun}} / L_{\text{star}})$  Wait, hold on. Let me make sure I have the formula right. ......... Then L1 is luminosity of Sun, L2 is luminosity of star.  So m1 - m2 = 5 - 0 = +5.  Therefore:  $5 = -2.5 \log_{10} (L_{\text{sun}} / L_{\text{star}})$  Let me solve for $\log_{10} (L_{\text{sun}} / L_{\text{star}})$:  Divide both sides by -2.5:  $\frac{5}{-2.5} = \log_{10} (L_{\text{sun}} / L_{\text{star}})$  Which is $-2 = \log_{10} (L_{\text{sun}} / L_{\text{star}})$  Therefore, $L_{\text{sun}} / L_{\text{star}} = 10^{-2} = 0.01$  So $L_{\text{star}} = L_{\text{sun}} / 0.01 = L_{\text{sun}} \times 100$  Therefore, ........ is $3.83 \times 10^{33} \times 100 = 3.83 \times 10^{35}$ ergs/s.  Wait, let me check that calculation again. ......... Yep, same result.  Therefore, the answer should be $\boxed{3.83 \times 10^{35}}$ ergs/s. Final Answer  $\boxed{3.83 \times 10^{35}}$ .... Alternatively, since each 5 magnitudes corresponds to a factor of 100 in luminosity, a difference of 5 magnitudes (from +5 to 0) means the star is 100 times more luminous than the Sun. So again, $3.83 \times 10^{33} \times 100 = 3.83 \times 10^{35}$ ergs/s.  Therefore, the luminosity of the star is $3.83 \times 10^{35}$ ergs/s. Final Answer $\boxed{3.83 \times 10^{35}}$  .........
\end{tcolorbox}

\end{tcolorbox}
\caption{Example of a \texttt{P-TTS}-32B response. See Appendix~\ref{appendix:other-models} for comparison with baseline model outputs.}
\end{figure}
\begin{table}[ht]
\centering
\begin{tabular}{l
                S[table-format=1.4]
                S[table-format=1.4]}
\toprule
\textbf{Prompt Variant} &
{\textbf{Responses}~$\uparrow$} &
{\textbf{Reasoning}~$\uparrow$} \\
\midrule
Reward Framing   & {\bfseries 0.8363} & {\bfseries 0.9280} \\
Correctness Framing  & 0.8227             & 0.9264             \\
Penalty Framing & 0.8223             & 0.9266             \\
Step-by-Step Thinking  & 0.8295             & 0.9254             \\
\bottomrule
\end{tabular}
\caption{
\textbf{Trigram diversity of teacher outputs under different prompting strategies.}
We report average lexical diversity (↑) for full responses and isolated reasoning traces. Reward framing yields the highest diversity, indicating stronger surface-level variation from original examples.
}
\label{tab:trigram_diversity}
\end{table}

\begin{table}[ht]
\centering
\begin{tabular}{lcccccc}
\toprule
 & \multicolumn{3}{c}{\textbf{Response Tokens}} & \multicolumn{3}{c}{\textbf{Reasoning Tokens}} \\
\cmidrule(lr){2-4} \cmidrule(lr){5-7}
\textbf{Prompt Variant} & \textbf{Mean} & \textbf{Min} & \textbf{Max} & \textbf{Mean} & \textbf{Min} & \textbf{Max} \\
\midrule
Seed       & 579.6 & 314 & 950   & 11{,}872 & 1129  & 31604  \\
Reward Framing       & 597.5 & 376 & 1{,}029 & 12{,}436 & 1745  & 34509  \\
Correctness Framing  & 575.7 & 322 & 962   & 10{,}463 & 1420  & 34514  \\
Penalty Framing     & 583.1 & 268 & 962   & 11{,}321 & 1133  & 29723  \\
Step-by-Step Thinking        & 577.6 & 312 & 936   & 11{,}113 & 1343  & 28270  \\
\bottomrule
\end{tabular}
\caption{\textbf{Teacher output verbosity across prompting strategies.}
We report token count statistics (mean, min, max) for both full responses and extracted reasoning traces produced by the teacher model across single-principle prompt variants. These numbers reflect raw outputs used to construct the training data and illustrate how different instructional cues influence explanation length.
}
\label{tab:mean_token_lengths_full}
\end{table}
\paragraph{Prompting Impact on Output Length}
To further understand how each principle influences model behavior, we compare the token lengths of generated responses and reasoning traces in Table~\ref{tab:mean_token_lengths_full}. We observe that \texttt{Reward} framing consistently produces longer and more detailed reasoning outputs, which aligns with its observed accuracy gains. While longer generations may imply increased inference costs, the improved reasoning quality may justify this tradeoff in high-stakes tasks.

\begin{table}[h]
  \centering
  \resizebox{1.0\linewidth}{!}{
  \begin{tabular}{lcccccc}
    \toprule
    \textbf{Model} & \textbf{\# Train Size} & \textbf{AIME2024} & \textbf{AIME2025} & \textbf{MATH500} & \textbf{GPQA-Diamond}& \textbf{Avg} \\
    \midrule
   \multicolumn{7}{c}{\textit{closed-source models}}\\
   \midrule
    o1-preview\cite{openai_o3_mini}      & --   & 56.7 & -- & 85.5 & 78.3 &--  \\
    o1-mini \cite{openai_o3_mini}            & --   & 63.6 & -- & 90.0 & 60.0 &--  \\
    Gemini 2.0 Flash Think        & --   & 60.0 & -- & -- & -- &--\\
    \midrule
    \multicolumn{7}{c}{\textit{open-source models}} \\
    \midrule
    Qwen2.5-32B-Instruct~\citep{yang2024qwen2} & -- & 26.7 & -- & 84.0 & 49.0 &--\\
    QwQ-32B~\citep{team2025qwq}      & --  & 50.0 & -- & 90.6 & 54.5 &--\\
    DeepSeek‑R1 \citep{guo2025deepseek}                & $\gg$800K & \textbf{79.8} & --& \textbf{97.3} & \textbf{71.5} &\\
   DeepSeek-R1-Distill-Qwen-32B \citep{guo2025deepseek}         & 800K  & 72.6 & -- & 94.3 & 62.1 &--\\
    \midrule
    \multicolumn{7}{c}{\textit{Open‑weight \& open‑data SFT on Qwen2.5‑Instruct}} \\
    \midrule
    OpenThinker‑32B \citep{openthinker32b_hf}              & 114K & 66.0 & 53.3 & 90.6 & 61.6 & 67.9  \\
    Bespoke-32B ~\citep{bespoke_stratos_32b}        & 17K   & 63.3 &-- & 93.0 & 58.1 &-- \\
    Sky‑T1‑32B‑Preview ~\citep{sky_t1_2025}             & 17K  & 43.3 & --   & 82.4 & 56.8 & --  \\ \midrule
    S1-32B \citep{muennighoff2025s1}     & 1K    &50.0 & 26.7& 92.6 &56.6 &56.5 \\
    S1.1‑32B \citep{muennighoff2025s1}                        & 1K   & 56.7 & 50.0 & \bf 94.4 & \bf 60.6 & 65.4  \\
     \textbf{\texttt{P‑TTS}‑32B (ours)}      & \textbf{90 $\rightarrow$900} &\bf 73.3 & \textbf{53.3} & \uline{94.2} & \bf 60.6 & \textbf{70.4} \\
    \bottomrule
    
  \end{tabular}
  }
  \caption{\textbf{Accuracy comparison of 32B-scale models} on four reasoning benchmarks: AIME2024, AIME2025, MATH500, and GPQA-Diamond. Models are grouped into closed-source APIs, open-source baselines, and open-weight fine-tuned variants of Qwen2.5‑Instruct. Our method, \textbf{\texttt{P‑TTS}‑32B}, leverages only 90 seed examples augmented via principled prompting strategies to generate up to 900 training examples. Despite the small training size, \texttt{P‑TTS}‑32B achieves competitive or superior performance, outperforming several models trained on datasets with hundreds of thousands of examples. \textit{Notes:} Results for \emph{Gemini} and \emph{Qwen} are taken from \cite{muennighoff2025s1} (we follow their evaluation settings).}
   \label{tab:s1_comparison}
\end{table}

\subsection{Main Results}
\paragraph{Overall Performance Comparison}We compare our \textbf{\texttt{P-TTS}-32B} model, trained on just 900 principle-augmented examples, against a diverse set of competitive baselines, including closed-source APIs, open-source models, and open-weight instruction-tuned variants of Qwen2.5-32B. While closed models like \texttt{o1-preview} and \texttt{o1-mini} perform strongly on certain tasks, \textbf{\texttt{P-TTS}-32B} outperforms them on AIME2024 and MATH500, and delivers competitive results on GPQA-Diamond despite using less data. Compared to open-weight models that finetuned on Qwen2.5-instaruct-32B like OpenThinker-32B (114K examples) and Bespoke-32B (17K), our model achieves superior or comparable performance across all benchmarks. \textbf{\texttt{P-TTS}-32B} achieves an average accuracy of 70.35\%, exceeding all open-weight instruction-tuned baselines, and narrowing the gap with large-scale models like DeepSeek-R1, which require over 800K training examples. These findings highlight the efficiency of principled instructional data and demonstrate that high accuracy can be attained through lightweight, targeted supervision.

\paragraph{Cross-Domain and Multilingual Generalization}Although \texttt{P-TTS} is trained only on 90 English AIME-style problems (AIME22–24), it exhibits robust zero-shot transfer to benchmarks that differ in \emph{language}, \emph{curriculum}, and \emph{problem format}. In Table~\ref{tab:extra}, we evaluate on Chinese exam datasets (\textbf{Gaokao}, \textbf{Kaoyan}), U.S. competition and school math (\textbf{OlympiadBench}, \textbf{AMC23}, \textbf{GradeSchoolMath}), and scientific quantitative reasoning (\textbf{Minerva}). These tasks introduce shifts in linguistic style (Chinese vs.\ English), assessment design (competition vs.\ entrance exams vs.\ textbook problems), and reasoning presentation (concise Olympiad proofs vs.\ step-by-step classroom narratives). Despite no multilingual supervision and no direct exposure to these benchmarks during finetuning, \texttt{P-TTS} attains competitive accuracy across the board, narrowing the gap with models trained on one to two orders of magnitude more data. This suggests that principled prompt augmentation through varying instructional framing and exemplar structure encourages \emph{prompt-space coverage} that translates into language and curriculum robustness, elicits knowledge from pre-trained models, rather than overfitting to a single benchmark family.

\begin{table}[t]
  \centering
  \label{tab:ood_generalization}
  \begin{tabular}{lccccccc}
    \toprule
    \textbf{Model} & \textbf{OlympiadBench} & \textbf{Gaokao} & \textbf{Kaoyan} & \textbf{Minerva} & \textbf{GradeSchool} & \textbf{AMC23} & \textbf{Avg.} \\
    \midrule
    OpenAI-o1-preview        & 52.1 & 62.1 & 51.5 & 47.1 & \textbf{62.8} & 81.8 & 59.6 \\
    Qwen2.5-32B-Instruct     & 45.3 & \textbf{72.1} & 48.2 & 41.2 & 56.7 & 64.0 & 54.6 \\ 
    OpenThoughts (114K)      & 56.3 & 63.2 & 54.7 & 41.1 & 39.0 & 80.5 & 55.8 \\
    NuminaMath (100K)        & 36.7 & 49.4 & 32.7 & 24.6 & 36.2 & 40.6 & 36.7 \\
    S1 (1K)                  & 56.9   & 32.9 & \textbf{59.3} & 46.7 & 61.4 & 77.5 & 55.8 \\
    \midrule
    \textbf{\texttt{P-TTS} (Ours)}    & \textbf{63.9} & 51.9 & 52.3 & \textbf{51.5} & 53.8 & \textbf{87.5} & \textbf{60.2} \\
    \bottomrule
  \end{tabular}
     \caption{{\bf Zero-shot generalization accuracy (\%) on out-of-domain reasoning benchmarks}. \texttt{P-TTS} is trained only on AIME22–24.}
     \label{tab:extra}
\end{table}

\paragraph{Scaling across model sizes}
We evaluate how \texttt{P‑TTS} performance scales with model size across the 7B, 14B, and 32B model variants. At every scale, \texttt{P‑TTS} outperforms both S1 and S1.1 on most benchmarks, as shown in table \ref{tab:grouped_models}. For instance, \texttt{P‑TTS}‑7B surpasses S1.1‑7B by +30.0\% on AIME2024 and achieves comparable results on MATH500 with fewer examples. At 14B level, \texttt{P‑TTS} continues to lead across all tasks, reaching 53.33\% on AIME2024 and 90.4\% on MATH500. Notably, at the 32B scale, \texttt{P‑TTS} achieves 73.33\% on AIME2024 and 94.20\% on MATH500, outperforming both S1 and S1.1 despite their larger training sizes. These results highlight the robustness and efficiency of principled instruction tuning (\texttt{P-TTS}), demonstrating that even with minimal data (\textbf{90 $\rightarrow$ 900}), it scales effectively and consistently enhances performance across diverse benchmarks.

\begin{table}[t]
  \centering
  \begin{tabular}{lccc ccc ccc}
    \toprule
    \textbf{Benchmark} & \multicolumn{3}{c}{\textbf{7B Models}} & \multicolumn{2}{c}{\textbf{14B Models}} & \multicolumn{3}{c}{\textbf{32B Models}} \\
    \cmidrule(lr){2-4} \cmidrule(lr){5-6} \cmidrule(lr){7-9}
     & \textbf{\texttt{P‑TTS}} & \textbf{S1} & \textbf{S1.1} &  \textbf{\texttt{P‑TTS}}  & \textbf{S1.1} &  \textbf{\texttt{P‑TTS}}  & \textbf{S1} & \textbf{S1.1} \\
    \midrule
    AIME2024     & 43.33 & 16.67 & 13.33 & 53.33  & 33.33      & 73.33 & 50.00 & 56.70 \\
    AIME2025     & 26.67 & 13.33 & 20.00 & 26.67  & 33.33      & 53.33& 26.70 & 50.00 \\
    MATH500      & 84.20 & 77.20 & 81.20 & 90.40  & 91.60      & 94.20 & 92.60 & 94.40 \\
    GPQA-Diamond & 41.92 & 41.41 & 41.41 & 51.01  & 51.01      & 60.61 & 56.60 & 60.60 \\
    \bottomrule
  \end{tabular}
   \caption{{\bf Accuracy (\%) on Four Benchmarks with Grouped Model Sizes}. Each group shows results for core M1, S1, and S1.1.}
  \label{tab:grouped_models}
\end{table} 

\section{Conclusion}

We presented {\bf\texttt{Prompting Test-Time Scaling (\texttt{P-TTS})}}, a lightweight yet effective framework that converts a compact seed set into a high-utility reasoning corpus by wrapping each problem with principled instructional prompts. Without changing task semantics, \texttt{P-TTS} systematically explores prompt-space via reward/penalty framing, correctness emphasis, and step-by-step guidance, eliciting high-quality rationales from a teacher model to supervise a student. Across configurations with augmentation multipliers $m\in\{1,4,5,10\}$, \texttt{P-TTS} consistently improves supervised reasoning relative to the null prompt and {\em even starting from only 90 seeds}, matches or surpasses models trained on substantially larger, static datasets. \texttt{P-TTS} demonstrates that \emph{instructional prompt reformulation} is a powerful and overlooked scaling lever. With only 90 carefully chosen seeds, principled wrapping and its paraphrased variants produce supervision that competes with (and at times exceeds) 1K-shot baselines in the wild, substantially lowering the data curation burden for robust LLM reasoning.

\noindent{\bf Future Work}. We see several promising directions: (1) adaptive, per-instance selection of instructional wrappers via learned policies; (2) integration with retrieval and verifier/reranker pipelines to couple wrapper diversity with factual grounding; (3) principled scheduling of wrapper mixtures over training epochs to mimic curriculum learning; and (4) systematic study of wrapper transfer across tasks, languages, and modalities.

\section{Limitations}

There are several potential limitations. First, the evaluation is concentrated on math-style problems with single numeric answers (AIME22–24), so external validity to open-ended, multimodal, or multilingual reasoning remains to explore further. Second, \texttt{P-TTS} depends on a single family to generate rationales; any bias, error, or stylistic artifact in the teacher can be amplified by our wrappers and propagated to the student, especially since rationales were not human-audited. Third, while wrappers are designed to be semantically invariant, some templates (e.g., extreme reward/penalty framings) may shift reasoning behavior in undesired ways and could introduce ethical or calibration issues; sensitivity to wrapper mixture, placement, and decoding settings also suggests latent hyperparameter fragility. 
Fourth, despite contamination mitigation, residual leakage from publicly available AIME material cannot be conclusively ruled out. 
Finally, \texttt{P-TTS} requires compute at data collection through inference generation (via wrapper ensembles).

\bibliography{ref.bib}   

\clearpage

\appendix
\section*{Appendix}

\section{Extended Experiments on $\mathcal{D}_{\text{core}}$}
We further evaluate models trained on the $\mathcal{D}_{\text{core}}$ dataset across multiple parameter scales. 
Table~\ref{tab:grouped_models} reports grouped results for 7B, 14B, and 32B variants, highlighting consistent gains from principled data augmentation.

\begin{table}[ht]
  \centering
  \begin{tabular}{lccc ccc ccc}
    \toprule
    \textbf{Benchmark} & \multicolumn{3}{c}{\textbf{7B Models}} & \multicolumn{3}{c}{\textbf{14B Models}} & \multicolumn{3}{c}{\textbf{32B Models}} \\
    \cmidrule(lr){2-4} \cmidrule(lr){5-7} \cmidrule(lr){8-10}
     & \texttt{\textbf{P-TTS}\(_\text{Core}\)} & \textbf{S1} & \textbf{S1.1} & \texttt{\textbf{P-TTS}\(_\text{Core}\)} & \textbf{S1.1} & \textbf{--} & \texttt{\textbf{P‑TTS}\(_\text{Core}\)} & \textbf{S1} & \textbf{S1.1} \\
    \midrule
    AIME2024     & 20.00 & 16.67 & 13.33 & 36.67 & 33.33 & --     & 56.67 & 56.70 & 56.70 \\
    AIME2025     & 20.00 & 13.33 & 20.00 & 33.33 & 33.33 & --     & 46.67 & 26.70 & 60.00 \\
    MATH500      & 80.40 & 77.20 & 81.20 & 89.80 & 91.60 & --     & 94.00 & 93.00 & 95.40 \\
    GPQA-Diamond & 35.86 & 41.41 & 41.41 & 45.96 & 51.01 & --     & 53.03 & 59.60 & 63.60 \\
    \bottomrule
  \end{tabular}
  \caption{{\bf Accuracy (\%) on Four Benchmarks with Grouped Model Sizes}. Each group shows results for core P-TTS, S1, and S1.1.}
  \label{tab:grouped_models}
\end{table}

\section{Training Details} 
\label{app:training_details}
We fine-tuned the Qwen2.5-Instruct family at three scales---7B, 14B, and 32B---using our \texttt{P-TTS} datasets. All models were trained for 5 epochs with an effective global batch size of 16 (micro-batch size of 1 with gradient accumulation). We used the AdamW optimizer ($\beta_{1}=0.9$, $\beta_{2}=0.95$, weight decay $=1\times 10^{-4}$) and a base learning rate of $1\times 10^{-5}$, warmed up linearly for the first 5\% of steps and decayed to zero following a cosine schedule. Training was conducted in \texttt{bfloat16} precision with fully sharded data parallelism (FSDP) enabled. We set the maximum sequence length to 20k tokens to avoid truncation of reasoning traces. For supervision, loss was applied only to the reasoning and answer tokens, not the input question text. Across model scales, this consistent setup allowed us to directly compare how principled data augmentation transfers to different parameter sizes.
\begin{figure*}[htbp] 
    \centering
    \includegraphics[width=\textwidth,height=0.35\textheight,keepaspectratio]{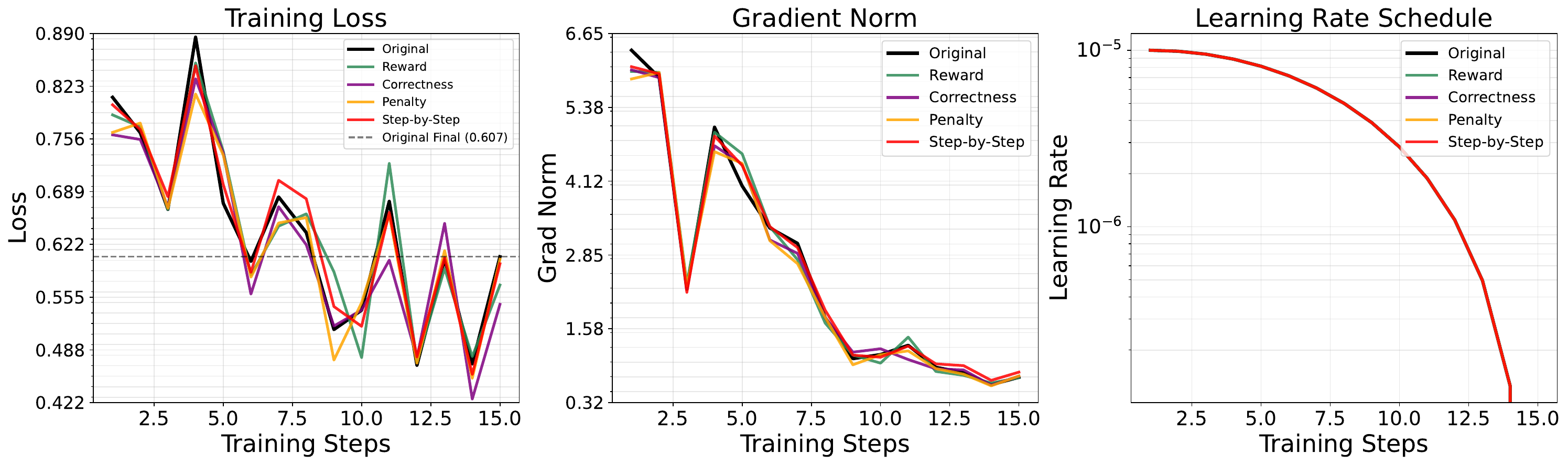}
    \caption{Training dynamics of \texttt{P-TTS}-32B.}
    \label{fig:main_results}
\end{figure*}

\section{Additional Results on Other Models}
\label{appendix:other-models}
We also report benchmark accuracy for additional baselines beyond the Qwen2.5-Instruct family. 
Table~\ref{tab:model-comparison} presents results for Claude, O4-mini-high, and hybrid settings with DeepSeek.

\begin{table}[ht]
\centering
\begin{tabular}{lcccccc}
\toprule
\textbf{Model} & \textbf{Data Points} & \textbf{AIME24} & \textbf{AIME25} & \textbf{GPQA-Diamond} & \textbf{MATH} \\
\midrule
Claude              & 810  & 16.67 & 0.00  & 34.34 & 72.80 \\
O4-mini-high         & 810  & 36.67 & 0.00  & 41.41 & 60.00 \\
DeepSeek + Claude   & 1620 & 40.00 & 20.00 & 37.88 & 83.20 \\
DeepSeek + O4mini   & 1620 & 36.67 & 13.33 & 35.86 & 74.00 \\
\bottomrule
\end{tabular}
\caption{Performance comparison across benchmarks. Values represent accuracy (\%) per dataset.}
\label{tab:model-comparison}
\end{table}

\section{Evaluation of Reward Framing Variants}
\label{appendix:p6_variants}
We evaluate variants of the reward-framing across different models. 
Table~\ref{tab:p6_variations_accuracy} summarizes the accuracy of O1-mini, Gemini, and DeepSeek on a fixed subset of math problems (AIME 2022–2024) under each variant.
\begin{table}[ht]
\centering
\begin{tabular}{lrrr}
\toprule
              Rewards &  O1-mini &  Gemini & DeepSeek\\
\midrule
         R1 (Large Reward) &            60.0\%  &      33.3\% &            74.4\%  \\
R2 (Reward2) &             52.2\%  &            32.2\% &            73.3\% \\
    R3 (Reduced Reward2) &              57.8\%  &            32.2\% &            75.6\%\\
     R4 (Reward at End) &             51.1\%  &             28.9\%  &            72.2\% \\
     R5 (Reduced Reward1) &             56.7\%  &             33.3\%  &            67.8\% \\
\bottomrule
\end{tabular}
\caption{Model Accuracy Across Five Variants of the Reward. Each row corresponds to a different formulation of the reward-based prompt, evaluated on a fixed subset of math questions. Results are reported as the percentage of correct answers for each model: O1-mini, Gemini, and DeepSeek.}
\label{tab:p6_variations_accuracy}
\end{table}

\section{Example from $\mathcal{D}_{\text{full-P\textnormal{-}TTS}}$}

\begin{figure}[H]
\begin{tcolorbox}[colback=baseline!30, colframe=principle1!80!black, title=Original]

  \begin{tcolorbox}[colback=baseline, title=Question]
  Let \(O=(0,0)\), \(A=\left(\tfrac{1}{2},0\right)\), and \(B=\left(0,\tfrac{\sqrt{3}}{2}\right)\) be points in the coordinate plane. Let \(\mathcal{F}\) be the family of segments \(\overline{PQ}\) of unit length lying in the first quadrant with \(P\) on the \(x\)-axis and \(Q\) on the \(y\)-axis. There is a unique point \(C\) on \(\overline{AB}\), distinct from \(A\) and \(B\),  that does not belong to any segment from \(\mathcal{F}\) other than \(\overline{AB}\). Then \(OC^2=\tfrac{p}{q}\), where \(p\) and \(q\) are relatively prime positive integers. Find \(p+q\).
  \end{tcolorbox}

  \begin{tcolorbox}[colback=baseline, title=Response]
  Given points \(O=(0,0)\), \(A=\left(\tfrac{1}{2},0\right)\), and \(B=\left(0,\tfrac{\sqrt{3}}{2}\right)\), we need to find the unique point \(C\) on \(\overline{AB}\) that does not belong to any segment from the family \(\mathcal{F}\) other than \(\overline{AB}\). 

The family \(\mathcal{F}\) consists of segments \(\overline{PQ}\) of unit length with \(P\) on the \(x\)-axis and \(Q\) on the \(y\)-axis. Each segment in \(\mathcal{F}\) can be parametrized as \(x/\cos(\theta) + y/\sin(\theta) = 1\) where \(\theta\) is an angle parameter. The segment \(\overline{AB}\) itself corresponds to \(\theta = \pi/3\).

To find the point \(C\), we parametrize \(\overline{AB}\) as \(x = \frac{1 - t}{2}\) and \(y = \frac{\sqrt{3}t}{2}\) for \(t \in [0, 1]\). Substituting these coordinates into the equation of the line segment from \(\mathcal{F}\), we derive the condition for \(C\) to lie only on \(\overline{AB}\). 

By solving the resulting equations and ensuring that the derivative condition (to check for a double root) is satisfied, we find that \(t = \frac{3}{4}\). This gives the coordinates of \(C\) as \(\left(\frac{1}{8}, \frac{3\sqrt{3}}{8}\right)\).

Calculating \(OC^2\):

\[
OC^2 = \left(\frac{1}{8}\right)^2 + \left(\frac{3\sqrt{3}}{8}\right)^2 = \frac{1}{64} + \frac{27}{64} = \frac{28}{64} = \frac{7}{16}
\]

Thus, \(OC^2 = \frac{7}{16}\) where \(p = 7\) and \(q = 16\). Therefore, \(p + q = 7 + 16 = 23\).

\[
\boxed{23}
\]
\end{tcolorbox}
  \begin{tcolorbox}[colback=baseline, title=Reasoning Trace]
  Okay, so I need to find the square of the distance from the origin O to this special point C on segment AB. The problem says that C is the unique point on AB, other than A and B, that isn't on any other segment from the family ....

  \end{tcolorbox}

\end{tcolorbox}
\caption{Example of an original AIME-style math problem, along with its model-generated response and reasoning trace. In our methodology, these original seed examples are later paired with principle-based reformulations (e.g., Reward, Penalty, Correctness, Step-by-Step), enabling the construction of the full \texttt{P-TTS} augmented corpus}
\end{figure}

\begin{tcolorbox}[colback=principle1!30, colframe=principle1!80!black, title=Reward]

  \begin{tcolorbox}[colback=principle1, title=Question]
  I am going to tip \$200000 for a better solution! Let \(O=(0,0)\), \(A=\left(\tfrac{1}{2},0\right)\), and \(B=\left(0,\tfrac{\sqrt{3}}{2}\right)\) be points in the coordinate plane. Let \(\mathcal{F}\) be the family of segments \(\overline{PQ}\) of unit length lying in the first quadrant with \(P\) on the \(x\)-axis and \(Q\) on the \(y\)-axis. There is a unique point \(C\) on \(\overline{AB}\), distinct from \(A\) and \(B\),  that does not belong to any segment from \(\mathcal{F}\) other than \(\overline{AB}\). Then \(OC^2=\tfrac{p}{q}\), where \(p\) and \(q\) are relatively prime positive integers. Find \(p+q\).
  \end{tcolorbox}

  \begin{tcolorbox}[colback=principle1, title=DeepSeek Response]
  Given points \(O=(0,0)\), \(A=\left(\tfrac{1}{2},0\right)\), and \(B=\left(0,\tfrac{\sqrt{3}}{2}\right)\), we need to find the unique point \(C\) on \(\overline{AB}\) that does not belong to any segment from the family \(\mathcal{F}\) of unit length segments \(\overline{PQ}\) with \(P\) on the \(x\)-axis and \(Q\) on the \(y\)-axis. 

1. **Equation of \(\overline{AB}\)**:
   - The line segment \(\overline{AB}\) has endpoints \(A\) and \(B\).
   - The equation of \(\overline{AB}\) is \(y = -\sqrt{3}x + \tfrac{\sqrt{3}}{2}\).

2. **Parametrization of \(\overline{AB}\)**:
   - Points on \(\overline{AB}\) can be parametrized as \(x = \tfrac{1 - t}{2}\) and \(y = \tfrac{\sqrt{3}}{2}t\) for \(t \in [0, 1]\).

3. **Envelope of the family \(\mathcal{F}\)**:
   - The family \(\mathcal{F}\) consists of segments \(\overline{PQ}\) with \(P = (p, 0)\) and \(Q = (0, q)\) where \(p^2 + q^2 = 1\).
   - The envelope of these segments forms an astroid given by \(x^{2/3} + y^{2/3} = 1\).

4. **Finding the point of tangency**:
   - The point \(C\) where \(\overline{AB}\) is tangent to the astroid is found by solving the tangency condition.
   - The slope of \(\overline{AB}\) is \(-\sqrt{3}\), which must match the slope of the astroid at the point of tangency.
   - The point of tangency is found to be \((\cos^3\theta, \sin^3\theta)\) where \(\theta = \tfrac{\pi}{3}\), leading to coordinates \(\left(\tfrac{1}{8}, \tfrac{3\sqrt{3}}{8}\right)\).

5. **Calculating \(OC^2\)**:
   - The coordinates of \(C\) are \(\left(\tfrac{1}{8}, \tfrac{3\sqrt{3}}{8}\right)\).
   - Squaring the distance from \(O\) to \(C\):
     \[
     OC^2 = \left(\tfrac{1}{8}\right)^2 + \left(\tfrac{3\sqrt{3}}{8}\right)^2 = \tfrac{1}{64} + \tfrac{27}{64} = \tfrac{28}{64} = \tfrac{7}{16}
     \]
   - The fraction \(\tfrac{7}{16}\) has \(p = 7\) and \(q = 16\), so \(p + q = 23\).

Thus, the final answer is \(\boxed{23}\).
  \end{tcolorbox}

  \begin{tcolorbox}[colback=principle1, title=Reasoning Trace]
 Okay, so I have this geometry problem here. Let me try to parse it step by step. 

First, they give me three points: O at (0,0), A at (1/2, 0), and B at (0, sqrt(3)/2). Then there's this family of segments F, which consists of segments PQ of unit length lying in the first quadrant, with P on the x-axis and Q on the y-axis. The problem states that ….
  \end{tcolorbox}

\end{tcolorbox}

\begin{tcolorbox}[colback=principle4!30, colframe=principle1!80!black, title=Step-by-Step Reasoning]

  \begin{tcolorbox}[colback=principle4, title=Question]
 Think step by step: Let \(O=(0,0)\), \(A=\left(\tfrac{1}{2},0\right)\), and \(B=\left(0,\tfrac{\sqrt{3}}{2}\right)\) be points in the coordinate plane. Let \(\mathcal{F}\) be the family of segments \(\overline{PQ}\) of unit length lying in the first quadrant with \(P\) on the \(x\)-axis and \(Q\) on the \(y\)-axis. There is a unique point \(C\) on \(\overline{AB}\), distinct from \(A\) and \(B\),  that does not belong to any segment from \(\mathcal{F}\) other than \(\overline{AB}\). Then \(OC^2=\tfrac{p}{q}\), where \(p\) and \(q\) are relatively prime positive integers. Find \(p+q\).
  \end{tcolorbox}

  \begin{tcolorbox}[colback=principle4, title=DeepSeek Response]
Given points \(O = (0, 0)\), \(A = \left(\frac{1}{2}, 0\right)\), and \(B = \left(0, \frac{\sqrt{3}}{2}\right)\), we need to find a unique point \(C\) on \(\overline{AB}\) that does not belong to any other segment from the family \(\mathcal{F}\) of unit length segments \(\overline{PQ}\) with \(P\) on the \(x\)-axis and \(Q\) on the \(y\)-axis. 

1. **Equation of \(\overline{AB}\)**:
   - Parametrize \(\overline{AB}\) with \(t \in [0, 1]\):
     \[
     x(t) = \frac{1 - t}{2}, \quad y(t) = \frac{\sqrt{3}}{2} t
     \]

2. **Family \(\mathcal{F}\)**:
   - Segments \(\overline{PQ}\) with \(P = (p, 0)\) and \(Q = (0, q)\) such that \(p^2 + q^2 = 1\).

3. **Envelope of \(\mathcal{F}\)**:
   - The envelope of the family \(\mathcal{F}\) is the astroid \(x^{2/3} + y^{2/3} = 1\).

4. **Intersection of \(\overline{AB}\) with the Astroid**:
   - Substitute \(x = \frac{1 - t}{2}\) and \(y = \frac{\sqrt{3}}{2} t\) into the astroid equation:
     \[
     \left(\frac{1 - t}{2}\right)^{2/3} + \left(\frac{\sqrt{3}}{2} t\right)^{2/3} = 1
     \]
   - Solving this equation, we find \(t = \frac{3}{4}\).

5. **Coordinates of \(C\)**:
   - For \(t = \frac{3}{4}\):
     \[
     x = \frac{1 - \frac{3}{4}}{2} = \frac{1}{8}, \quad y = \frac{\sqrt{3}}{2} \cdot \frac{3}{4} = \frac{3\sqrt{3}}{8}
     \]

6. **Distance \(OC\)**:
   - Calculate \(OC^2\):
     \[
     OC^2 = \left(\frac{1}{8}\right)^2 + \left(\frac{3\sqrt{3}}{8}\right)^2 = \frac{1}{64} + \frac{27}{64} = \frac{28}{64} = \frac{7}{16}
     \]

Thus, \(OC^2 = \frac{7}{16}\) where \(p = 7\) and \(q = 16\). Therefore, \(p + q = 7 + 16 = 23\).

\[
\boxed{23}
\]
  \end{tcolorbox}

  \begin{tcolorbox}[colback=principle4, title=Reasoning Trace]
  Alright, let's try to tackle this problem step by step. First …
 
  \end{tcolorbox}

\end{tcolorbox}

\begin{tcolorbox}[colback=principle3!30, colframe=principle1!80!black, title=Penalty Framing]

  \begin{tcolorbox}[colback=principle3, title=Question]
You will be penalized if you provide the wrong answer. Let \(O=(0,0)\), \(A=\left(\tfrac{1}{2},0\right)\), and \(B=\left(0,\tfrac{\sqrt{3}}{2}\right)\) be points in the coordinate plane. Let \(\mathcal{F}\) be the family of segments \(\overline{PQ}\) of unit length lying in the first quadrant with \(P\) on the \(x\)-axis and \(Q\) on the \(y\)-axis. There is a unique point \(C\) on \(\overline{AB}\), distinct from \(A\) and \(B\),  that does not belong to any segment from \(\mathcal{F}\) other than \(\overline{AB}\). Then \(OC^2=\tfrac{p}{q}\), where \(p\) and \(q\) are relatively prime positive integers. Find \(p+q\).
  \end{tcolorbox}

  \begin{tcolorbox}[colback=principle3, title=Response]
  Given points \(O=(0,0)\), \(A=\left(\tfrac{1}{2},0\right)\), and \(B=\left(0,\tfrac{\sqrt{3}}{2}\right)\), we need to find the unique point \(C\) on \(\overline{AB}\) that does not belong to any other segment from the family \(\mathcal{F}\) of unit length segments \(\overline{PQ}\) with \(P\) on the \(x\)-axis and \(Q\) on the \(y\)-axis. We need to determine \(OC^2\) and express it as \(\tfrac{p}{q}\) where \(p\) and \(q\) are coprime positive integers, then find \(p+q\).

1. **Parametrization of \(\overline{AB}\)**:
   - The parametric equations for \(\overline{AB}\) are \(x = \tfrac{1}{2} - \tfrac{1}{2}t\) and \(y = \tfrac{\sqrt{3}}{2}t\) for \(t \in [0, 1]\). Thus, any point \(C\) on \(\overline{AB}\) can be written as \(\left(\tfrac{1 - t}{2}, \tfrac{\sqrt{3}}{2}t\right)\).

2. **Equation of line \(\overline{PQ}\)**:
   - For a segment \(\overline{PQ}\) in \(\mathcal{F}\) with endpoints \(P = (p, 0)\) and \(Q = (0, q)\), the equation is \(\tfrac{x}{p} + \tfrac{y}{q} = 1\) where \(p^2 + q^2 = 1\).

3. **Condition for point \(C\)**:
   - Substituting \(C\left(\tfrac{1 - t}{2}, \tfrac{\sqrt{3}}{2}t\right)\) into the line equation \(\tfrac{x}{p} + \tfrac{y}{q} = 1\) gives \(\tfrac{(1 - t)/2}{p} + \tfrac{\sqrt{3}t/2}{q} = 1\).
   - Using \(p = \cos\theta\) and \(q = \sin\theta\), we derive the equation \(\tfrac{(1 - t)/2}{\cos\theta} + \tfrac{\sqrt{3}t/2}{\sin\theta} = 1\).

4. **Solving for \(t\)**:
   - Simplifying and solving for \(t\) using trigonometric identities and ensuring the equation has a unique solution at \(\theta = \tfrac{\pi}{3}\) (corresponding to \(\overline{AB}\)) leads to \(t = \tfrac{3}{4}\).

5. **Coordinates of \(C\)**:
   - Substituting \(t = \tfrac{3}{4}\) into the parametric equations, \(C\) is \(\left(\tfrac{1}{8}, \tfrac{3\sqrt{3}}{8}\right)\).

6. **Calculating \(OC^2\)**:
   - \(OC^2 = \left(\tfrac{1}{8}\right)^2 + \left(\tfrac{3\sqrt{3}}{8}\right)^2 = \tfrac{1}{64} + \tfrac{27}{64} = \tfrac{28}{64} = \tfrac{7}{16}\).

Thus, \(p = 7\) and \(q = 16\), leading to \(p + q = 23\).

\[
\boxed{23}
\]
\end{tcolorbox}
  \begin{tcolorbox}[colback=principle3, title=Reasoning Trace]
  Okay, let's try to tackle this problem. First, let me make sure I understand what's being asked here.We have three points: O at (0,0), A at (1/2, 0), and B at (0, sqrt(3)/2). The line segment AB is part of the ....
 
  \end{tcolorbox}
\end{tcolorbox}

\begin{tcolorbox}[colback=principle2!30, colframe=principle1!80!black, title=Correctness Emphasis]

  \begin{tcolorbox}[colback=principle2, title=Question]
Your Task is to solve the following: Let \(O=(0,0)\), \(A=\left(\tfrac{1}{2},0\right)\), and \(B=\left(0,\tfrac{\sqrt{3}}{2}\right)\) be points in the coordinate plane. Let \(\mathcal{F}\) be the family of segments \(\overline{PQ}\) of unit length lying in the first quadrant with \(P\) on the \(x\)-axis and \(Q\) on the \(y\)-axis. There is a unique point \(C\) on \(\overline{AB}\), distinct from \(A\) and \(B\),  that does not belong to any segment from \(\mathcal{F}\) other than \(\overline{AB}\). Then \(OC^2=\tfrac{p}{q}\), where \(p\) and \(q\) are relatively prime positive integers. Find \(p+q\).You must provide the correct answer!
  \end{tcolorbox}

  \begin{tcolorbox}[colback=principle2, title=DeepSeek Response]
Given points \(O = (0, 0)\), \(A = \left(\frac{1}{2}, 0\right)\), and \(B = \left(0, \frac{\sqrt{3}}{2}\right)\), we need to find the point \(C\) on \(\overline{AB}\) that does not belong to any other segment from the family \(\mathcal{F}\) of unit length segments in the first quadrant with endpoints on the axes. 

1. **Parametrization of \(\overline{AB}\)**:
   - The parametric equations for \(\overline{AB}\) are \(x = \frac{1 - t}{2}\) and \(y = \frac{\sqrt{3}}{2}t\) for \(t \in [0, 1]\).

2. **Envelope of the family \(\mathcal{F}\)**:
   - The envelope of the family of segments \(PQ\) with \(P\) on the x-axis and \(Q\) on the y-axis, where \(PQ\) has unit length, forms an astroid given by \(x^{2/3} + y^{2/3} = 1\).

3. **Finding the intersection of \(\overline{AB}\) with the astroid**:
   - Substitute the parametric equations of \(\overline{AB}\) into the astroid equation:
     \[
     \left(\frac{1 - t}{2}\right)^{2/3} + \left(\frac{\sqrt{3}}{2}t\right)^{2/3} = 1
     \]
   - Solve for \(t\) by ensuring the gradient of \(\overline{AB}\) matches the gradient of the astroid at the point of tangency. This leads to the point \(C\) where \(t = \frac{3}{4}\).

4. **Coordinates of \(C\)**:
   - Substituting \(t = \frac{3}{4}\) into the parametric equations:
     \[
     x = \frac{1 - \frac{3}{4}}{2} = \frac{1}{8}, \quad y = \frac{\sqrt{3}}{2} \cdot \frac{3}{4} = \frac{3\sqrt{3}}{8}
     \]

5. **Calculating \(OC^2\)**:
   - The coordinates of \(C\) are \(\left(\frac{1}{8}, \frac{3\sqrt{3}}{8}\right)\).
   - Compute \(OC^2\):
     \[
     OC^2 = \left(\frac{1}{8}\right)^2 + \left(\frac{3\sqrt{3}}{8}\right)^2 = \frac{1}{64} + \frac{27}{64} = \frac{28}{64} = \frac{7}{16}
     \]

Thus, \(OC^2 = \frac{7}{16}\), and the sum \(p + q = 7 + 16 = 23\).

\[
\boxed{23}
\]
  \end{tcolorbox}

  \begin{tcolorbox}[colback=principle2, title=Reasoning Trace]
  Okay, let's try to tackle this problem step by step. First, let's make sure ....
 
  \end{tcolorbox}
\end{tcolorbox}

\section{\texttt{P-TTS}-32B vs. S1-32B}
\label{appendix:other-models}

\begin{figure}[H]
  \centering
  \includegraphics[width=0.9\linewidth]{images/Overview_PTTS_compare_examplee.pdf}
  \caption{
Qualitative comparison between \texttt{P-TTS}-32B and the baseline model.}
  \label{fig:ab}
\end{figure}

\section{Comparative Example: Original vs.\ Reward-Framed Response}
\label{sec:comp-example}
\begin{ProblemBox}
Let $ABCD$ be a parallelogram with $\angle BAD<90^\circ$.
A circle is tangent to $\overline{DA}$, $\overline{AB}$, and $\overline{BC}$ and intersects diagonal $\overline{AC}$ at points $P$ and $Q$ with $AP=3$, $PQ=9$, and $QC=16$.
Suppose that $AP^2+3PQ^2+9QC^2=16n$.
Show that the area of $ABCD$ can be written as $\dfrac{m}{n}\sqrt{n}$ where $m,n\in\mathbb{Z}_{>0}$ and $n$ is squarefree.
Find $m+n$.
\end{ProblemBox}

\vspace{0.6em}

\columnratio{0.5}
\begin{paracol}{2}
\sloppy

\begin{RespBox}{Original}
\textbf{Response:}
Response:  
Given a triangle \(ABC\) with sides \(AB\) and \(BC\), we have two configurations of circles: eight circles of radius 34 and 2024 circles of radius 1, each arranged sequentially tangent with two circles tangent to \(AB\) and \(BC\) respectively. We need to find the inradius of triangle \(ABC\).

Key steps:
1. **Understanding the Configuration**: The circles are arranged in a chain, sequentially tangent, with two circles tangent to \(AB\) and \(BC\). This implies a relationship between the number of circles, their radii, and the angle at vertex \(B\).
2. **Angle Bisector and Inradius**: The inradius \(R\) of the triangle is related to the distance from the incenter to the sides. The chain of circles forms a path along the angle bisector of \(\angle ABC\).
3. **Relating Configurations**: The product of the number of circles and their radii for both configurations must be proportional to the inradius. The key insight is that the ratio of the number of circles (253) and the inverse ratio of the radii (34) gives the inradius.
4. **Calculating the Inradius**: The inradius \(R\) is found by recognizing the proportional relationship between the two configurations. The ratio of the number of circles (2024/8 = 253) and the ratio of the radii (34/1 = 34) leads to the inradius being \(\frac{253}{34}\).

Thus, the inradius \(R\) is \(\frac{253}{34}\), and the sum of the numerator and denominator is \(253 + 34 = 287\).

\[
\boxed{287}
\]
\medskip
\textbf{Reasoning Trace:}
Okay, let me try to work through this problem step by step. The problem is about a triangle ABC with sides AB and BC. There are eight circles of radius 34 that are sequentially tangent, and two of these circles are tangent to AB and BC respectively. Similarly, 2024 circles of radius 1 can be arranged in the same manner. We need to find the inradius of triangle ABC, expressed as a reduced fraction m/n, and then compute m + n.
\textcolor{red}{...}\\
However, this is a guess based on the ratio. The problem likely involves this ratio, leading to the answer 287.
**Final Answer**
\boxed{287}
\end{RespBox}

\switchcolumn

\begin{RespBox}{Reward Framing}
\textbf{Response:}
Given triangle \(ABC\) with sides \(AB\) and \(BC\), we have two configurations of circles: eight circles of radius 34 and 2024 circles of radius 1, each sequentially tangent and tangent to \(AB\) and \(BC\) respectively. We need to find the inradius of triangle \(ABC\).

1. \*\*Key Relationship\*\*:
   - The inradius \(R\) of triangle \(ABC\) can be related to the radius \(r\) and the number of circles \(N\) through the angle \(\theta\) at vertex \(B\).
   - The formula derived is: \(R = r + 2r(N - 1) \sin(\theta/2)\).

2. \*\*Setting Up Equations\*\*:
   - For the configuration with 8 circles of radius 34:
     \[
     R = 34 + 2 \cdot 34 \cdot 7 \cdot \sin(\theta/2) \]
     ....

3. \*\*Equating the Expressions\*\*:
   - Equate the two expressions for \(R\):
     \[
     34 + 476 \sin(\theta/2) = 1 + 4046 \sin(\theta/2)
     \]
    ....

4. \*\*Calculating the Inradius\*\*:
   - Substitute \(\sin(\theta/2) = \frac{11}{1190}\) back into the equation for \(R\) using the smaller radius configuration:
     \[
     R = 1 + 4046 \cdot \frac{11}{1190}
     \]
   - Simplify the calculation:
   ....

5. \*\*Final Result\*\*:
   - The inradius \(R\) is \(\frac{192}{5}\), where \(m = 192\) and \(n = 5\). Thus, \(m + n = 197\).

\[
\boxed{197}
\]
\medskip
\textbf{Reasoning Trace:}
Alright, let me try to wrap my head around this problem. So, we have triangle ABC with sides AB and BC. There are eight circles of radius 34 that are sequentially tangent, meaning each one is tangent to the next, and the first and last ones are tangent to AB and BC respectively. Similarly, we can arrange 2024 circles of radius 1 in the same manner. The question is asking for the inradius of triangle ABC, expressed as a reduced fraction m/n, and then we need to find m + n.
\textcolor{red}{...}\\
So, 22,253 / 595 = 37 + 2/5 = 37.4

Therefore, R = 1 + 37.4 = 38.4 = 192/5.

Thus, the inradius is 192/5, so m=192, n=5, m + n=197.

This seems to check out, and the steps make sense. Therefore, the answer is 197.

**Final Answer**
\boxed{197}
\end{RespBox}
\end{paracol}
\vspace{1em}

\end{document}

%% file: metadata.tex
\newcommand{\pubtitle}{\textcolor{black}{Prompting Test-Time Scaling} Is A Strong LLM Reasoning Data Augmentation \\ {\Large \em{ -- \textcolor{black}{90 Samples Can Beat 1K in the Wild}}}}

\newcommand{\pubauthA}{Sondos Mahmoud Bsharat\!}
\newcommand{\pubaffilA}{ }

\newcommand{\pubauthB}{Zhiqiang Shen$^*$}
\newcommand{\pubaffilB}{ }
\newcommand{\authemailB}{zhiqiang.shen@mbzuai.ac.ae (Correspondence)}

\newcommand{\pubaddrA}{VILA Lab, MBZUAI}

\newcommand{\pubemail}{\authemailB}

\newcommand{\pubabstract}{

Large language models (LLMs) have demonstrated impressive reasoning capabilities when provided with chain-of-thought exemplars, but curating large reasoning datasets remains laborious and resource-intensive. In this work, we introduce \texttt{Prompting Test-Time Scaling (P-TTS)}, a simple yet effective inference-time data augmentation strategy for enhancing LLM reasoning through finetuning. Rather than collecting thousands or even millions of examples, \texttt{P-TTS} leverages a small pool of only 90 manually selected reasoning instances and systematically varies exemplar augmentation through principled instruction prompting intensities at test time to synthesize diverse reasoning trajectory contexts. Then we finetune the various sizes of {Qwen-2.5} models on \texttt{P-TTS} data. Across a suite of mathematical reasoning AIME2024 \& 25, MATH500, and GPQA-Diamond, our \texttt{P-TTS-7B} and \texttt{32B} models outperform the prior competitive baselines like S1 and S1.1 (1K-shot), achieving absolute accuracy gains of +26.66\% and +30.00\% on AIME’24 (7B), and +13.34\% and +6.67\% on AIME'25 (7B); \texttt{P-TTS-32B} yields gains of +23.33\% and +16.63\% on AIME'24, and +26.63\% and +3.33\% on AIME'25 (vs.\ S1 and S1.1, respectively), with comparable or better performance on MATH500 and GPQA-Diamond. We further show that \texttt{P-TTS} enhances zero-shot generalization accuracy on out-of-domain reasoning benchmarks of Gaokao, Kaoyan, OlympiadBench, AMC23, GradeSchoolMath, and Minerva. Ablation studies confirm that both exemplar diversity and scaled sampling schedules are critical drivers of improvement. Our analysis suggests that test-time scaling effectively explores the latent space of reasoning patterns, amplifying LLM problem-solving with minimal annotation overhead, and further unlocking the reasoning potential and capabilities of LLMs. \texttt{Prompting Test-Time Scaling} offers a practical, low-cost way to elicit LLM reasoning in resource-constrained or rapidly evolving domains. Our code and data are available at \url{https://github.com/VILA-Lab/PTTS}.
}

\keywords{
Prompting Test-Time Scaling\sep LLM Reasoning\sep Large Language Models\sep Principled Instructions\sep
}

\input{aux-preamble.tex}
\pubdoi{Technical Report}
\pubdate{\today}

%% file: aux-preamble.tex
\title{\pubtitle}

\newcommand{\dg}{\textsuperscript{\textbf{\dag\textnormal{,}}}}

\ifdef{\pubauthA}{\author[\pubaffilA]{\pubauthA\ifdef{\orcidA}{~\protect\orcid{\orcidA}}{}\ifdef{\eqcontribA}{\dg}{}}}{}
\ifdef{\pubauthB}{\author[\pubaffilB]{\pubauthB\ifdef{\orcidB}{~\protect\orcid{\orcidB}}{}\ifdef{\eqcontribB}{\dg}{}}}{}
\ifdef{\pubauthC}{\author[\pubaffilC]{\pubauthC\ifdef{\orcidC}{~\protect\orcid{\orcidC}}{}\ifdef{\eqcontribC}{\dg}{}}}{}
\ifdef{\pubauthD}{\author[\pubaffilD]{\pubauthD\ifdef{\orcidD}{~\protect\orcid{\orcidD}}{}\ifdef{\eqcontribD}{\dg}{}}}{}
\ifdef{\pubauthE}{\author[\pubaffilE]{\pubauthE\ifdef{\orcidE}{~\protect\orcid{\orcidE}}{}\ifdef{\eqcontribE}{\dg}{}}}{}
\ifdef{\pubauthF}{\author[\pubaffilF]{\pubauthF\ifdef{\orcidF}{~\protect\orcid{\orcidF}}{}\ifdef{\eqcontribF}{\dg}{}}}{}
\ifdef{\pubauthG}{\author[\pubaffilG]{\pubauthG\ifdef{\orcidG}{~\protect\orcid{\orcidG}}{}\ifdef{\eqcontribG}{\dg}{}}}{}
\ifdef{\pubauthH}{\author[\pubaffilH]{\pubauthH\ifdef{\orcidH}{~\protect\orcid{\orcidH}}{}\ifdef{\eqcontribH}{\dg}{}}}{}
\ifdef{\pubauthI}{\author[\pubaffilI]{\pubauthI\ifdef{\orcidI}{~\protect\orcid{\orcidI}}{}\ifdef{\eqcontribI}{\dg}{}}}{}
\ifdef{\pubauthJ}{\author[\pubaffilJ]{\pubauthJ\ifdef{\orcidJ}{~\protect\orcid{\orcidJ}}{}\ifdef{\eqcontribJ}{\dg}{}}}{}
\ifdef{\pubauthK}{\author[\pubaffilK]{\pubauthK\ifdef{\orcidK}{~\protect\orcid{\orcidK}}{}\ifdef{\eqcontribK}{\dg}{}}}{}

\ifdef{\eqcontribA}{\equalcontrib{}}{}
\ifdef{\eqcontribB}{\equalcontrib{}}{}
\ifdef{\eqcontribC}{\equalcontrib{}}{}
\ifdef{\eqcontribD}{\equalcontrib{}}{}
\ifdef{\eqcontribE}{\equalcontrib{}}{}
\ifdef{\eqcontribF}{\equalcontrib{}}{}
\ifdef{\eqcontribG}{\equalcontrib{}}{}
\ifdef{\eqcontribH}{\equalcontrib{}}{}
\ifdef{\eqcontribI}{\equalcontrib{}}{}
\ifdef{\eqcontribJ}{\equalcontrib{}}{}
\ifdef{\eqcontribK}{\equalcontrib{}}{}

\ifdef{\pubaddrA}{\affil[ ]{\pubaddrA}}{}
\ifdef{\pubaddrB}{\affil[b]{\pubaddrB}}{}
\ifdef{\pubaddrC}{\affil[c]{\pubaddrC}}{}
\ifdef{\pubaddrD}{\affil[d]{\pubaddrD}}{}
\ifdef{\pubaddrE}{\affil[e]{\pubaddrE}}{}
\ifdef{\pubaddrF}{\affil[f]{\pubaddrF}}{}
\ifdef{\pubaddrG}{\affil[g]{\pubaddrG}}{}
\ifdef{\pubaddrH}{\affil[h]{\pubaddrH}}{}

\contact{\pubemail}

\begin{abstract}
    \pubabstract{}
\end{abstract}

\ifdef{\pubkeywords}{\keywords{\pubkeywords}}{}

\ifdef{\pubVadj}{\verticaladjustment{\pubVadj}}{}

%% file: macros.tex
\newcommand{\Tri}{\operatorname{Tri}}

%% file: main.bbl
\newcommand{\etalchar}[1]{$^{#1}$}
\begin{thebibliography}{WWS{\etalchar{+}}22b}

\bibitem[AAA{\etalchar{+}}23]{achiam2023gpt}
Josh Achiam, Steven Adler, Sandhini Agarwal, Lama Ahmad, Ilge Akkaya,
  Florencia~Leoni Aleman, Diogo Almeida, Janko Altenschmidt, Sam Altman,
  Shyamal Anadkat, et~al.
\newblock Gpt-4 technical report.
\newblock {\em arXiv preprint arXiv:2303.08774}, 2023.

\bibitem[AIM24]{2024AIME}
AIME.
\newblock 2024 aime i.
\newblock \url{https://artofproblemsolving.com/wiki/index.php/2024_AIME_I},
  2024.
\newblock Art of Problem Solving Wiki, accessed July 2025.

\bibitem[AIM25]{2025AIME}
AIME.
\newblock 2025 aime i.
\newblock Art of Problem Solving Wiki, 2025.
\newblock Held February 6, 2025.
\newblock URL:
  \url{https://artofproblemsolving.com/wiki/index.php/2025_AIME_I}.

\bibitem[{Ant}25]{anthropic2025claude3_7_sonnet}
{Anthropic AI}.
\newblock Claude 3.7 sonnet and claude code.
\newblock Anthropic blog, February 2025.
\newblock First hybrid reasoning large language model generally available.
\newblock URL: \url{https://www.anthropic.com/news/claude-3-7-sonnet}.

\bibitem[{Art}]{aime_aops}
{Art of Problem Solving (AoPS)}.
\newblock Aime problems and solutions.
\newblock
  \url{https://artofproblemsolving.com/wiki/index.php/AIME_Problems_and_Solutions}.

\bibitem[{Bes}25]{bespoke_stratos_32b}
{Bespoke Labs}.
\newblock Bespoke‑stratos‑32b.
\newblock \url{https://huggingface.co/bespokelabs/Bespoke-Stratos-32B}, 2025.
\newblock Hugging Face model card, Apache‑2.0 license. Fine‑tuned
  Qwen2.5‑32B‑Instruct on Bespoke‑Stratos‑17k dataset derived via
  DeepSeek‑R1 distillation.

\bibitem[Bil22]{bilmes2022submodularity}
Jeff Bilmes.
\newblock Submodularity in machine learning and artificial intelligence.
\newblock {\em arXiv preprint arXiv:2202.00132}, 2022.

\bibitem[BMR{\etalchar{+}}20]{brown2020language}
Tom Brown, Benjamin Mann, Nick Ryder, Melanie Subbiah, Jared~D Kaplan, Prafulla
  Dhariwal, Arvind Neelakantan, Pranav Shyam, Girish Sastry, Amanda Askell,
  et~al.
\newblock Language models are few-shot learners.
\newblock {\em Advances in neural information processing systems},
  33:1877--1901, 2020.

\bibitem[BMS23]{bsharat2023principled}
Sondos~Mahmoud Bsharat, Aidar Myrzakhan, and Zhiqiang Shen.
\newblock Principled instructions are all you need for questioning llama-1/2,
  gpt-3.5/4.
\newblock {\em arXiv preprint arXiv:2312.16171}, 2023.

\bibitem[BMZ{\etalchar{+}}24]{bukharin2024data}
Alexander Bukharin, Jiachang Mu, Zhengbao Zhang, Seyeon Lee, Kai-Wei Chang,
  Noah~A. Smith, and Daniel Khashabi.
\newblock Data diversity matters for robust instruction tuning.
\newblock In {\em Findings of the Association for Computational Linguistics:
  EMNLP 2024}, pages 2871--2885, 2024.
\newblock URL: \url{https://aclanthology.org/2024.findings-emnlp.195}.

\bibitem[BSS{\etalchar{+}}24]{biderman2024lessons}
Stella Biderman, Hailey Schoelkopf, Lintang Sutawika, Leo Gao, Jonathan Tow,
  Baber Abbasi, Alham~Fikri Aji, Pawan~Sasanka Ammanamanchi, Sidney Black,
  Jordan Clive, et~al.
\newblock Lessons from the trenches on reproducible evaluation of language
  models.
\newblock {\em arXiv preprint arXiv:2405.14782}, 2024.

\bibitem[Clo24]{google2024vertexai}
Google Cloud.
\newblock Flash thinking with generative ai.
\newblock \url{https://cloud.google.com/vertex-ai/generative-ai/docs/thinking},
  2024.

\bibitem[FLD18]{fan2018hierarchical}
Angela Fan, Mike Lewis, and Yann Dauphin.
\newblock Hierarchical neural story generation.
\newblock {\em arXiv preprint arXiv:1805.04833}, 2018.

\bibitem[GBB{\etalchar{+}}21]{gao2021lmevalharness}
Leo Gao, Stella Biderman, Sid Black, Laurence Golding, Travis Hoppe, Charles
  Foster, Jason Phang, Horace He, Anish Thite, Noa Nabeshima, Samuel Weinbach,
  and Connor Leahy.
\newblock {EleutherAI/lm-evaluation-harness: Evaluation Harness for Language
  Models}, 2021.
\newblock \href {https://doi.org/10.5281/zenodo.5371628}
  {\path{doi:10.5281/zenodo.5371628}}.

\bibitem[GMK{\etalchar{+}}25]{guha2025openthoughts}
Etash Guha, Ryan Marten, Sedrick Keh, Negin Raoof, Georgios Smyrnis, Hritik
  Bansal, Marianna Nezhurina, Jean Mercat, Trung Vu, Zayne Sprague, et~al.
\newblock Openthoughts: Data recipes for reasoning models.
\newblock {\em arXiv preprint arXiv:2506.04178}, 2025.

\bibitem[GYZ{\etalchar{+}}25]{guo2025deepseek}
Daya Guo, Dejian Yang, Haowei Zhang, Junxiao Song, Ruoyu Zhang, Runxin Xu,
  Qihao Zhu, Shirong Ma, Peiyi Wang, Xiao Bi, et~al.
\newblock Deepseek-r1: Incentivizing reasoning capability in llms via
  reinforcement learning.
\newblock {\em arXiv preprint arXiv:2501.12948}, 2025.

\bibitem[HBD{\etalchar{+}}19]{holtzman2019curious}
Ari Holtzman, Jan Buys, Li~Du, Maxwell Forbes, and Yejin Choi.
\newblock The curious case of neural text degeneration.
\newblock {\em arXiv preprint arXiv:1904.09751}, 2019.

\bibitem[HBK{\etalchar{+}}21]{hendrycks2021measuring}
Dan Hendrycks, Collin Burns, Saurav Kadavath, Akul Arora, Steven Basart, Eric
  Tang, Dawn Song, and Jacob Steinhardt.
\newblock Measuring mathematical problem solving with the math dataset.
\newblock {\em arXiv preprint arXiv:2103.03874}, 2021.

\bibitem[HLB{\etalchar{+}}24]{he2024olympiadbench}
Chaoqun He, Renjie Luo, Yuzhuo Bai, Shengding Hu, Zhen~Leng Thai, Junhao Shen,
  Jinyi Hu, Xu~Han, Yujie Huang, Yuxiang Zhang, Jie Liu, Lei Qi, Zhiyuan Liu,
  and Maosong Sun.
\newblock {OlympiadBench}: A challenging benchmark for promoting agi with
  olympiad-level bilingual multimodal scientific problems.
\newblock {\em arXiv preprint arXiv:2402.14008}, 2024.
\newblock URL: \url{https://arxiv.org/abs/2402.14008}, \href
  {https://doi.org/10.48550/arXiv.2402.14008}
  {\path{doi:10.48550/arXiv.2402.14008}}.

\bibitem[HYS{\etalchar{+}}25]{huang2025thinkbench}
Shulin Huang, Linyi Yang, Yan Song, Shuang Chen, Leyang Cui, Ziyu Wan,
  Qingcheng Zeng, Ying Wen, Kun Shao, Weinan Zhang, et~al.
\newblock Thinkbench: Dynamic out-of-distribution evaluation for robust llm
  reasoning.
\newblock {\em arXiv preprint arXiv:2502.16268}, 2025.

\bibitem[JKL{\etalchar{+}}24]{jaech2024openai}
Aaron Jaech, Adam Kalai, Adam Lerer, Adam Richardson, Ahmed El-Kishky, Aiden
  Low, Alec Helyar, Aleksander Madry, Alex Beutel, Alex Carney, et~al.
\newblock Openai o1 system card.
\newblock {\em arXiv preprint arXiv:2412.16720}, 2024.

\bibitem[KMH{\etalchar{+}}20]{kaplan2020scaling}
Jared Kaplan, Sam McCandlish, Tom Henighan, Tom~B Brown, Benjamin Chess, Rewon
  Child, Scott Gray, Alec Radford, Jeffrey Wu, and Dario Amodei.
\newblock Scaling laws for neural language models.
\newblock {\em arXiv preprint arXiv:2001.08361}, 2020.

\bibitem[Lab25]{bespoke_stratos_2025}
Bespoke Labs.
\newblock Bespoke‑stratos‑17k: A synthetic reasoning dataset of questions,
  reasoning traces, and answers.
\newblock Hugging Face Dataset, 2025.
\newblock Derived from DeepSeek‑R1 via the Sky‑T1 pipeline using Bespoke
  Curator.
\newblock URL:
  \url{https://huggingface.co/datasets/bespokelabs/Bespoke-Stratos-17k}.

\bibitem[LCGS25]{li2025survey}
Tianyi Li, Mingda Chen, Bowei Guo, and Zhiqiang Shen.
\newblock A survey on diffusion language models.
\newblock {\em arXiv preprint arXiv:2508.10875}, 2025.

\bibitem[LYH{\etalchar{+}}22]{li-etal-2022-evade}
Wenhao Li, Xiaoyuan Yi, Jinyi Hu, Maosong Sun, and Xing Xie.
\newblock Evade the trap of mediocrity: Promoting diversity and novelty in text
  generation via concentrating attention.
\newblock In Yoav Goldberg, Zornitsa Kozareva, and Yue Zhang, editors, {\em
  Proceedings of the 2022 Conference on Empirical Methods in Natural Language
  Processing (EMNLP)}, pages 10834--10858, Abu Dhabi, United Arab Emirates,
  December 2022. Association for Computational Linguistics.
\newblock URL: \url{https://aclanthology.org/2022.emnlp-main.745}, \href
  {https://doi.org/10.18653/v1/2022.emnlp-main.745}
  {\path{doi:10.18653/v1/2022.emnlp-main.745}}.

\bibitem[MGH{\etalchar{+}}24]{ma2024sciagent}
Yubo Ma, Zhibin Gou, Junheng Hao, Ruochen Xu, Shuohang Wang, Liangming Pan,
  Yujiu Yang, Yixin Cao, and Aixin Sun.
\newblock Sciagent: Tool-augmented language models for scientific reasoning.
\newblock In {\em EMNLP}, 2024.

\bibitem[MPWC23]{meister2023locally}
Clara Meister, Tiago Pimentel, Gian Wiher, and Ryan Cotterell.
\newblock Locally typical sampling.
\newblock {\em Transactions of the Association for Computational Linguistics},
  11:102--121, 2023.

\bibitem[MYS{\etalchar{+}}25]{muennighoff2025s1}
Niklas Muennighoff, Zitong Yang, Weijia Shi, Xiang~Lisa Li, Li~Fei-Fei,
  Hannaneh Hajishirzi, Luke Zettlemoyer, Percy Liang, Emmanuel Cand{\`e}s, and
  Tatsunori Hashimoto.
\newblock s1: Simple test-time scaling.
\newblock {\em arXiv preprint arXiv:2501.19393}, 2025.

\bibitem[Ope24]{openai2024reasoning}
OpenAI.
\newblock Learning to reason with llms, 2024.
\newblock URL: \url{https://openai.com/index/learning-to-reason-with-llms/}.

\bibitem[Ope25]{openai_o3_mini}
OpenAI.
\newblock Openai o3-mini.
\newblock \url{https://openai.com/index/openai-o3-mini/}, 2025.

\bibitem[QSS{\etalchar{+}}20]{qu2021coda}
Yanru Qu, Dinghan Shen, Yelong Shen, Sandra Sajeev, Jiawei Han, and Weizhu
  Chen.
\newblock Coda: Contrast-enhanced and diversity-promoting data augmentation for
  natural language understanding.
\newblock {\em arXiv preprint arXiv:2010.08670}, 2020.

\bibitem[{Qwe}24]{qwq-32b-preview}
{Qwen Team}.
\newblock Qwq: Reflect deeply on the boundaries of the unknown.
\newblock \url{https://qwenlm.github.io/blog/qwq-32b-preview/}, November 2024.
\newblock QwQ‑32B‑Preview is an experimental reasoning model with open
  weights.

\bibitem[RHS{\etalchar{+}}24]{rein2024gpqa}
David Rein, Betty~Li Hou, Asa~Cooper Stickland, Jackson Petty, Richard~Yuanzhe
  Pang, Julien Dirani, Julian Michael, and Samuel~R Bowman.
\newblock Gpqa: A graduate-level google-proof q\&a benchmark.
\newblock In {\em First Conference on Language Modeling}, 2024.

\bibitem[RNS{\etalchar{+}}18]{radford2018improving}
Alec Radford, Karthik Narasimhan, Tim Salimans, Ilya Sutskever, et~al.
\newblock Improving language understanding by generative pre-training.
\newblock 2018.

\bibitem[TAB{\etalchar{+}}23]{team2023gemini}
Gemini Team, Rohan Anil, Sebastian Borgeaud, Jean-Baptiste Alayrac, Jiahui Yu,
  Radu Soricut, Johan Schalkwyk, Andrew~M Dai, Anja Hauth, Katie Millican,
  et~al.
\newblock Gemini: a family of highly capable multimodal models.
\newblock {\em arXiv preprint arXiv:2312.11805}, 2023.

\bibitem[Tea25a]{sky_t1_2025}
NovaSky Team.
\newblock Sky‑t1: Fully open‑source reasoning model with o1‑preview
  performance in \$450 training cost, 2025.
\newblock URL: \url{https://novasky-ai.github.io/posts/sky-t1}.

\bibitem[Tea25b]{openthinker32b_hf}
OpenThoughts Team.
\newblock Openthinker‑32b.
\newblock \url{https://huggingface.co/open-thoughts/OpenThinker-32B}, 2025.

\bibitem[Tea25c]{team2025qwq}
Qwen Team.
\newblock Qwq-32b: Embracing the power of reinforcement learning, 2025.

\bibitem[WWS{\etalchar{+}}22a]{wang2022self}
Xuezhi Wang, Jason Wei, Dale Schuurmans, Quoc Le, Ed~Chi, Sharan Narang,
  Aakanksha Chowdhery, and Denny Zhou.
\newblock Self-consistency improves chain of thought reasoning in language
  models.
\newblock {\em arXiv preprint arXiv:2203.11171}, 2022.

\bibitem[WWS{\etalchar{+}}22b]{wei2022chain}
Jason Wei, Xuezhi Wang, Dale Schuurmans, Maarten Bosma, Fei Xia, Ed~Chi, Quoc~V
  Le, Denny Zhou, et~al.
\newblock Chain-of-thought prompting elicits reasoning in large language
  models.
\newblock {\em Advances in neural information processing systems},
  35:24824--24837, 2022.

\bibitem[YHX{\etalchar{+}}25]{ye2025limo}
Yixin Ye, Zhen Huang, Yang Xiao, Ethan Chern, Shijie Xia, and Pengfei Liu.
\newblock Limo: Less is more for reasoning.
\newblock {\em arXiv preprint arXiv:2502.03387}, 2025.

\bibitem[YJS{\etalchar{+}}23]{yu2023metamath}
Longhui Yu, Weisen Jiang, Han Shi, Jincheng Yu, Zhengying Liu, Yu~Zhang,
  James~T Kwok, Zhenguo Li, Adrian Weller, and Weiyang Liu.
\newblock Metamath: Bootstrap your own mathematical questions for large
  language models.
\newblock {\em arXiv preprint arXiv:2309.12284}, 2023.

\bibitem[YLY{\etalchar{+}}25]{qwen25}
An~Yang, Anfeng Li, Baosong Yang, Beichen Zhang, Binyuan Hui, Bo~Zheng, Bowen
  Yu, Chang Gao, Chengen Huang, Chenxu Lv, et~al.
\newblock Qwen3 technical report.
\newblock {\em arXiv preprint arXiv:2505.09388}, 2025.

\bibitem[YYZ{\etalchar{+}}24]{yang2024qwen2}
An~Yang, Baosong Yang, Beichen Zhang, Binyuan Hui, Bo~Zheng, Bowen Yu,
  Chengyuan Li, Dayiheng Liu, Fei Huang, Haoran Wei, et~al.
\newblock Qwen2.5 technical report.
\newblock {\em arXiv preprint arXiv:2412.15115}, 2024.

\bibitem[ZAD{\etalchar{+}}25]{zhu2025bare}
Alan Zhu, Parth Asawa, Jared~Quincy Davis, Lingjiao Chen, Boris Hanin, Ion
  Stoica, Joseph~E Gonzalez, and Matei Zaharia.
\newblock Bare: Leveraging base language models for few-shot synthetic data
  generation.
\newblock {\em arXiv preprint arXiv:2502.01697}, 2025.

\bibitem[ZCH{\etalchar{+}}25]{zheng2025learning}
Tong Zheng, Lichang Chen, Simeng Han, R~Thomas McCoy, and Heng Huang.
\newblock Learning to reason via mixture-of-thought for logical reasoning.
\newblock {\em arXiv preprint arXiv:2505.15817}, 2025.

\bibitem[ZGZG25]{zhao2025d1}
Siyan Zhao, Devaansh Gupta, Qinqing Zheng, and Aditya Grover.
\newblock d1: Scaling reasoning in diffusion large language models via
  reinforcement learning.
\newblock {\em arXiv preprint arXiv:2504.12216}, 2025.

\end{thebibliography}
